\documentclass[sigconf, 10pt]{acmart}

\usepackage{url}
\usepackage{graphicx}
\usepackage{algorithmic}
\usepackage{subfigure}
\usepackage{amsmath}
\usepackage{xcolor}
\usepackage{makecell}
\usepackage{booktabs}
\usepackage{multirow}
\usepackage{balance}
\usepackage{listings}
\usepackage{booktabs}
\usepackage{pifont}
\usepackage{enumitem}
\usepackage[ruled,vlined,linesnumbered]{algorithm2e}
\usepackage[inkscapelatex=false]{svg}
\usepackage{bm}
\usepackage{array}
\usepackage{ragged2e}
\usepackage{tabularray}
\usepackage[normalem]{ulem}
\usepackage{enumitem}

\acmConference[Submitted for review to ACM MobiCom'25]{}
\acmYear{}
\copyrightyear{}

\newcommand{\ie}{\textit{i}.\textit{e}.}
\newcommand{\eg}{\textit{e}.\textit{g}.}
\newcommand{\etc}{\textit{etc}}

\newcommand{\revise}[1]{\textcolor{red}{#1}}

\begin{document}
\begin{sloppypar}

\title{AutoIOT: LLM-Driven Automated Natural Language Programming for AIoT Applications}

\author{Leming Shen\textsuperscript{1},
    Qiang Yang\textsuperscript{2},
    Yuanqing Zheng\textsuperscript{1},
    Mo Li\textsuperscript{3}
}

\affiliation{
    \textsuperscript{1}The Hong Kong Polytechnic University,
    \textsuperscript{2}University of Cambridge,\\
    \textsuperscript{3}Hong Kong University of Science and Technology
    \country{}
}

\email{
    leming.shen@connect.polyu.hk, qy258@cam.ac.uk, csyqzheng@comp.polyu.edu.hk, lim@cse.ust.hk
}

\renewcommand{\shortauthors}{Leming Shen, Qiang Yang, Yuanqing Zheng, Mo Li}

\begin{abstract}

The advent of Large Language Models (LLMs) has profoundly transformed our lives, revolutionizing interactions with AI and lowering the barrier to AI usage. While LLMs are primarily designed for natural language interaction, the extensive embedded knowledge empowers them to comprehend digital sensor data. This capability enables LLMs to engage with the physical world through IoT sensors and actuators, performing a myriad of AIoT tasks. Consequently, this evolution triggers a paradigm shift in conventional AIoT application development, democratizing its accessibility to all by facilitating the design and development of AIoT applications via natural language. However, some limitations need to be addressed to unlock the full potential of LLMs in AIoT application development. First, existing solutions often require transferring raw sensor data to LLM servers, which raises privacy concerns, incurs high query fees, and is limited by token size. Moreover, the reasoning processes of LLMs are opaque to users, making it difficult to verify the robustness and correctness of inference results. This paper introduces \textit{AutoIOT}, an LLM-based automated program generator for AIoT applications. \textit{AutoIOT} enables users to specify their requirements using natural language (input) and automatically synthesizes interpretable programs with documentation (output). \textit{AutoIOT} automates the iterative optimization to enhance the quality of generated code with minimum user involvement. \textit{AutoIOT} not only makes the execution of AIoT tasks more explainable but also mitigates privacy concerns and reduces token costs with local execution of synthesized programs. Extensive experiments and user studies demonstrate \textit{AutoIOT}’s remarkable capability in program synthesis for various AIoT tasks. The synthesized programs can match and even outperform some representative baselines.
\end{abstract}

\vspace{-5pt}


\vspace{-5pt}

\acmYear{2025}\copyrightyear{2025}
\acmConference[MobiCom '25]{The 31st Annual International Conference on Mobile Computing and Networking}{Nov 4--8, 2025}{Hong Kong, China}
\acmBooktitle{The 31st Annual International Conference on Mobile Computing and Networking (ACM MobiCom ’25), Nov 4--8, 2025, Hong Kong, China}
\acmDOI{10.1145/xxxxxxx.xxxxx}
\acmISBN{nnn-n-nnnn-nnnn-n/nn/nn}

\maketitle
\vspace{-5pt}
\section{Introduction}

\vspace{-2pt}
Artificial Intelligence of Things (AIoT) \cite{lee2024mobilegpt, cai2023efficient, wen2024autodroid, huang2022real, shen2023feddm, liu2021mandipass} is an emerging paradigm that leverages advanced artificial intelligence (AI) algorithms to process a vast amount of data generated by Internet of Things (IoT) devices. This technology brings a new level of intelligence and automation to various applications, including healthcare \cite{ouyang2024admarker, ouyang2023harmony}, smart sensing \cite{yang2023aquahelper, yang2023voshield, yang2022deepear}, and autonomous driving \cite{he2023vi}. 



\begin{figure}[t]
\vspace{-12pt}
\setlength{\abovecaptionskip}{4pt}
    \centering
    \includegraphics[width=0.47\textwidth]{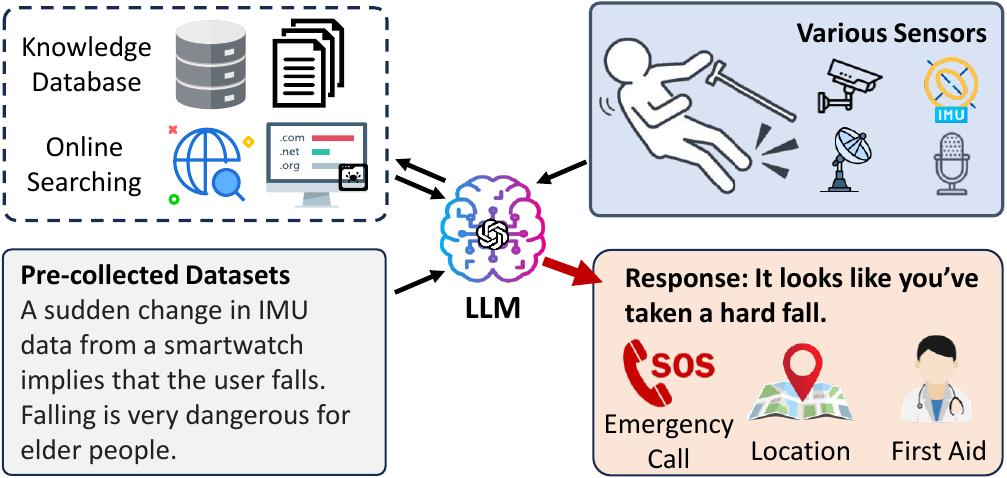}
    \caption{Illustration of how LLMs can sense and interact with the physical world in AIoT applications.}
    \label{fig:blueprint}
    \vspace{-18pt}
\end{figure}

Recent advances in large language models (LLMs) (\eg, GPT-4 \cite{achiam2023gpt}) fundamentally changed the way we interact with AI. While initially designed to understand natural languages, recent pioneering works \cite{10.1145/3638550.3641130, liu2023large, ji2024hargpt} have demonstrated considerable proficiency of LLMs in exploiting embedded world knowledge by interpreting IoT sensor data to perform various AIoT tasks. Recent works \cite{10.1145/3638550.3641130} term such an endeavor -- Penetrative AI. Fig.~\ref{fig:blueprint} illustrates how LLMs can be tasked to comprehend and even interact with the physical world through integration with IoT sensors and actuators.

However, current LLMs on AIoT tasks \cite{10.1145/3638550.3641130, liu2023large, quan2024sensorbench} fall short in supporting AIoT applications \cite{yu2024fdlora, yu2024revolutionizing, hou2024one, hou2023jamming, cao2022gaze, hou2025molora}: 1) The trustworthiness of the inference results is compromised since the inference process is performed inside a "black box" and opaque to users. Thus, the robustness of the applications or the correctness of the inference results are hard to verify; 2) Transmitting the raw or intermediate sensor data from user devices to LLM servers raises privacy concerns, incurs prohibitive query fees, and increases response latency; 3) Sensor data typically exhibits extensive length and high dimensionality, making remote processing at LLM servers infeasible due to token limits \cite{yang2024queueing, token_limit}. Ideally, the integration of LLMs with AIoT applications should be trustworthy, privacy-preserving, and communication-efficient.

On the other hand, existing works on LLMs have showcased their remarkable capabilities in code generation to accomplish a variety of programming tasks \cite{le2023codechain, jiang2023selfevolve, luo2023wizardcoder}. 
\textit{Can we leverage LLMs to synthesize programs to fulfill AIoT application requirements?}
This approach can 1) enhance the explainability and trustworthiness of the AIoT applications as the synthesized programs can be examined and interpreted by developers, 2) mitigate privacy concerns, and reduce the communication cost since the programs can be executed locally on user devices without offloading raw sensor data, and 3) efficiently process high-dimensional continuous sensor data without being limited by the token size or bounded by the round trip time over the network. To this end, we propose \textit{AutoIOT}, a user-friendly natural language programming system based on LLMs. \textit{AutoIOT} automatically identifies and retrieves the necessary domain knowledge over the Internet, intelligently synthesizes programs, and evolves the programs iteratively given sample inputs and ground truth. Surprisingly, we found that the synthesized programs can sometimes outperform some representative baselines and sample programs of recent academic papers.


While the automatic program synthesis for AIoT applications is promising and exciting, it entails tremendous technical challenges. 1) \textbf{High Complexity of AIoT Tasks.} Contrary to existing works that generate code for individual modules or well-defined functions \cite{austin2021program}, AIoT applications typically need a systematic design and integration involving multiple functional components, leading to much higher reasoning and planning complexity beyond the capability of current LLMs. To address this issue, \textit{AutoIOT} decomposes the programming task into several distinct modules and generates the corresponding code segments. In particular, \textit{AutoIOT} leverages chain-of-thought (CoT) prompts \cite{wei2022chain} to divide the task into a few sub-tasks and integrate their solutions, eventually making the sub-tasks manageable by LLMs. 2) \textbf{Lack of Domain Knowledge in AIoT.} LLMs are trained on pre-collected general corpus datasets, which may not include the latest domain-specific knowledge needed for the development of various emerging AIoT applications. 
To tackle this problem, \textit{AutoIOT} guides the LLMs to search and retrieve necessary knowledge and algorithms, thereby enabling in-context training and inference augmented with domain knowledge for LLMs. 3) \textbf{Heavy Intervention and Constant Feedback.} Our preliminary experiments (\S~\ref{sec:preliminary}) reveal that to generate functionally correct programs, developers have to give timely feedback to LLMs and constantly intervene in the entire development process. For example, developers need to provide specific reference materials and describe algorithms in great detail, which can be time-consuming and defeat the very purpose of automated natural language programming. Ideally,  \textit{AutoIOT} should be able to synthesize the program with no intervention from users and require minimum user input only when necessary. To this end, we develop \textit{AutoIOT} that can execute, debug, and optimize the synthesized program given sample inputs and outputs. 

We fully implement \textit{AutoIOT}\footnote{The project is available at \url{https://github.com/lemingshen/AutoIOT}} and evaluate its synthesized programs with four representative AIoT applications: heartbeat detection, IMU-based human activity recognition (HAR), mmWave-based HAR, and multimodal HAR. Extensive experiments and user studies show that, the synthesized programs can achieve comparable performance to the corresponding baselines and significant improvements in user satisfaction. Besides, \textit{AutoIOT} substantially reduces the communication cost and the total execution time. These findings demonstrate the LLM's exceptional proficiency and great potential in synthesizing programs for AIoT applications.
In summary, we make the following contributions:
\vspace{-2pt}
\begin{itemize}[leftmargin=8pt]
    \item To our best, \textit{AutoIOT} is the first work that enables systematic natural language programming for AIoT applications.
    \item We design and implement three novel technical modules (\ie, background knowledge retrieval module, CoT prompting-based program synthesis module, and automated code improvement module) to synthesize and optimize programs for AIoT applications.
    \item Our comprehensive experiments demonstrate that synthesized programs can achieve comparable performance to baselines and sometimes outperform them. 
    
\end{itemize}

\vspace{-8pt}
\section{Background \& Motivation}
\vspace{-2pt}
We first revisit the pioneering Penetrative AI efforts that leverage the embedded knowledge of LLMs to address AIoT tasks. Then, we present the results of our preliminary experiments to demonstrate the feasibility and identify the key challenges of implementing an LLM-driven automated natural language programming system for AIoT applications.

\begin{figure}[t]
\vspace{-13pt}
\setlength{\abovecaptionskip}{4pt}
    \centering
    \includegraphics[width=0.43\textwidth]{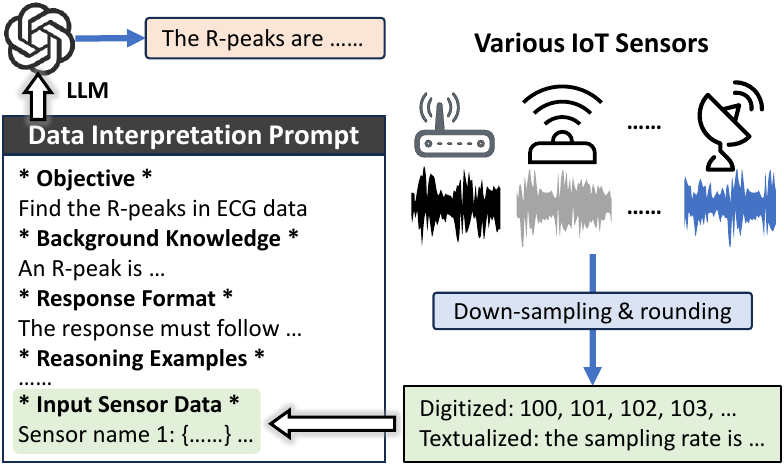}
    \caption{Prior work processes sensor data with LLMs.}
    \label{fig:penetrative_ai}
    \vspace{-18pt}
\end{figure}

\begin{figure}[t]
\vspace{0pt}
\setlength{\abovecaptionskip}{4pt}
    \centering
    \includegraphics[width=0.48\textwidth]{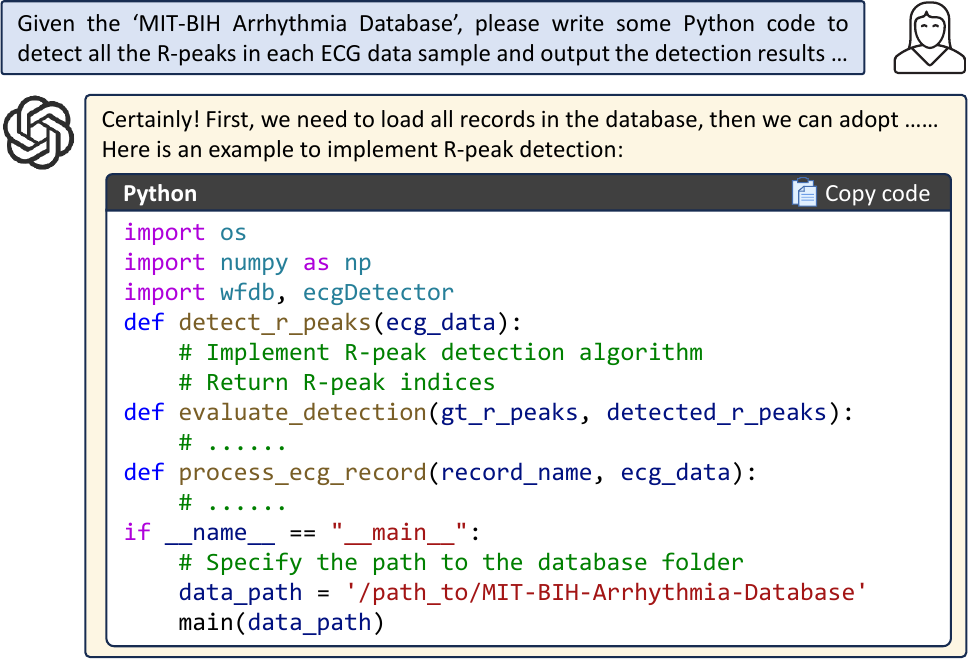}
    \caption{An example of direct code generation.}
    \label{fig:direct_code}
    \vspace{-18pt}
\end{figure}

\vspace{-8pt}
\subsection{Background}
\vspace{-2pt}

Fig.~\ref{fig:penetrative_ai} illustrates an example of recent work in Penetrative AI \cite{10.1145/3638550.3641130}, where various sensor data are textualized and embedded in a prompt, which is further used to instruct the LLM to perform inference tasks on the textualized sensor data.

Preliminary results indicate that the common knowledge embedded in LLMs can be leveraged to accomplish various real-world AIoT tasks (\eg, inferring a user's location via WiFi fingerprints, and counting heartbeats via raw ECG data). However, the inference processes of LLMs are largely opaque to users, rendering the results less explainable and trustworthy. Moreover, raw sensor data is transmitted from users to LLM servers, raising privacy concerns about sensor data.
Furthermore, limited by token size, existing work down-samples and quantizes raw sensor data, leading to degraded inference performance. The remote processing at LLM servers also necessitates the round-trip transmission of prompts and results over the network, which increases response latency. 

\begin{figure}[t]
\vspace{-12pt}
\setlength{\abovecaptionskip}{4pt}
    \centering
    \includegraphics[width=0.42\textwidth]{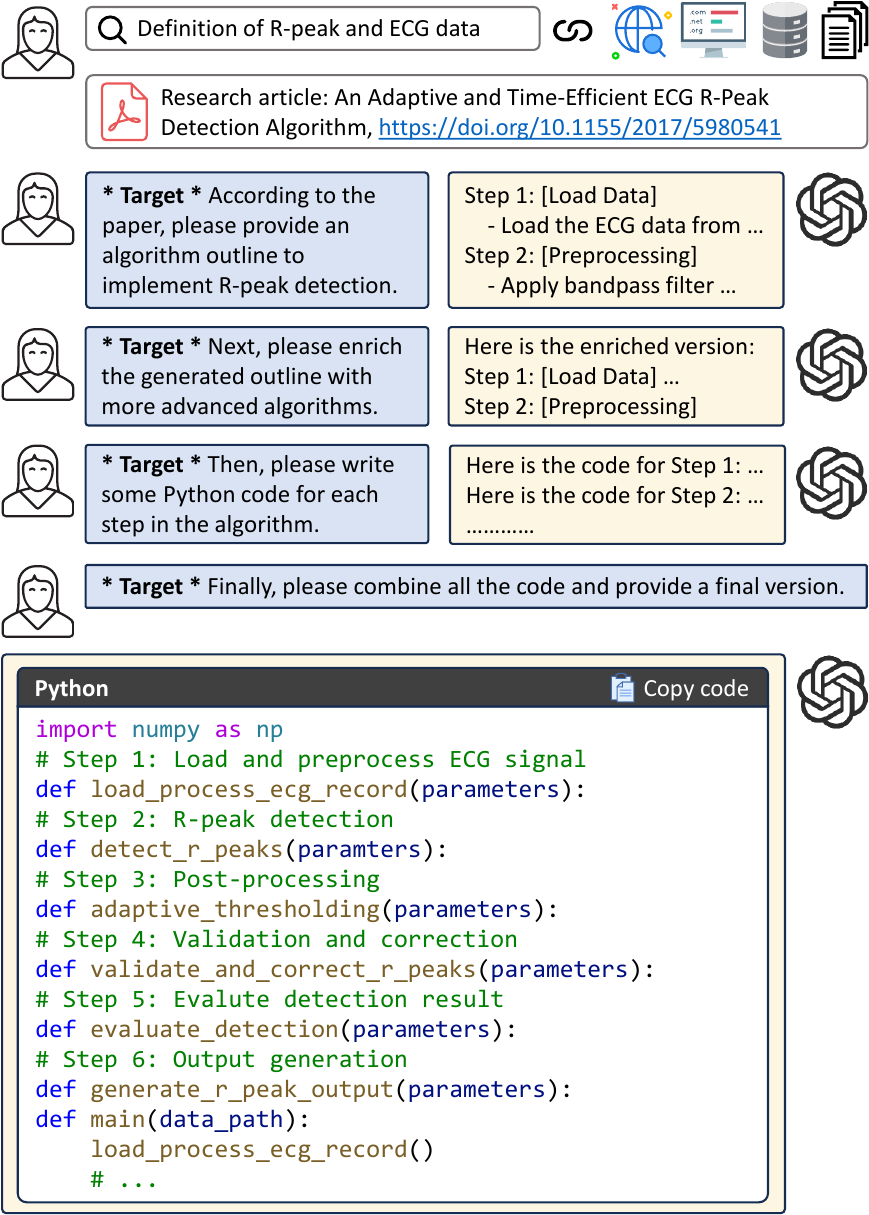}
    \caption{Code generation with user intervention.}
    \label{fig:user_intervention}
    \vspace{-18pt}
\end{figure}

To overcome these issues, we explore a new approach that leverages LLMs to synthesize AIoT programs and executes the programs locally to process users' data. This new approach allows developers to examine and verify the synthesized programs, protect data privacy, process sensor data streams without compromising data resolution or quality, and avoid transmission time over networks.



\vspace{-8pt}
\subsection{Preliminary Experiments}
\label{sec:preliminary}
\vspace{-2pt}
The latest LLMs \cite{du2024mercury, achiam2023gpt, team2023gemini} have demonstrated extraordinary proficiency in generating code snippets. For example, Mercury \cite{du2024mercury} leverages LLMs to generate code for well-defined programming tasks. \textit{Is it feasible to instruct LLMs to synthesize programs that can tackle AIoT tasks?} Our preliminary results show that it is possible yet extremely challenging for LLMs to synthesize functionally correct programs for AIoT tasks. Taking heartbeat detection as an example, as depicted in Fig.~\ref{fig:direct_code}. 
When we instruct the LLM to generate a program to process raw ECG waveform and detect heartbeats, the LLM can only generate a few null functions without concrete implementation or import some nonexistent packages.

\begin{figure*}[t]
\vspace{-12pt}
\setlength{\abovecaptionskip}{4pt}
    \centering
    \includegraphics[width=0.98\textwidth]{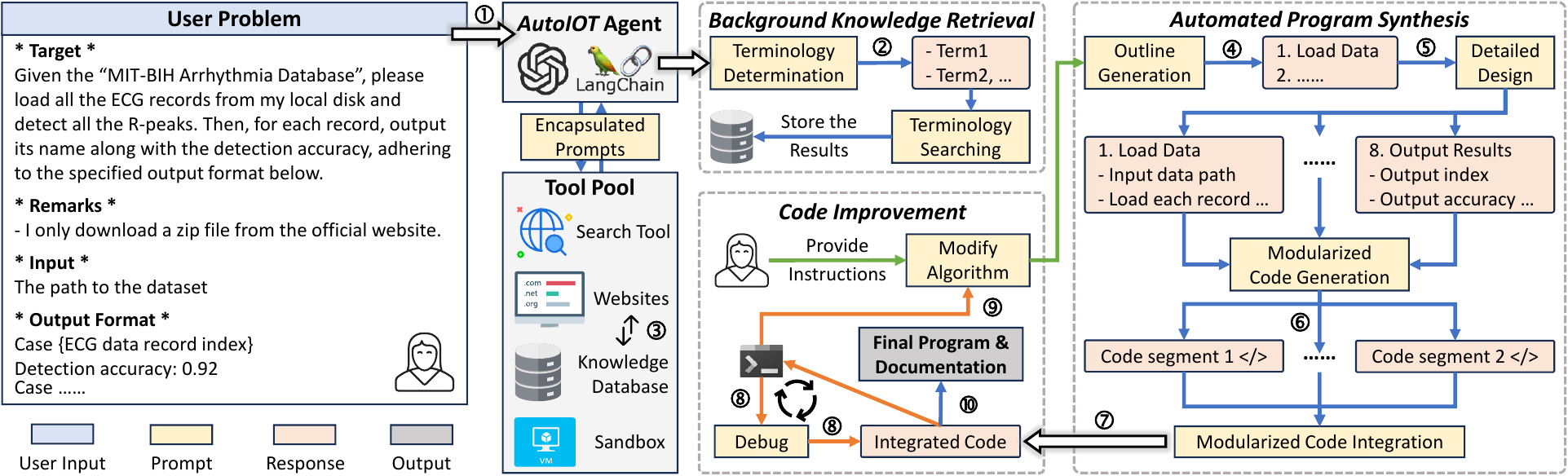}
    \caption{The system overview and workflow of \textit{AutoIOT}.}
    \label{fig:system_overview}
    \vspace{-12pt}
\end{figure*}

We hypothesize that the reasons for this might be threefold: 1) LLMs lack domain-specific knowledge, let alone the latest algorithms tailored for AIoT tasks. As a result, for highly specialized AIoT applications, LLMs can only offer some suggestions or generate high-level code outlines rather than detailed functional implementation. 2) AIoT applications typically require systematic programming, where multiple functional modules are first developed for different subtasks (\eg, signal preprocessing, data cleaning, neural network initialization), which are then constructively integrated to form a comprehensive and cohesive program. This development process involves much higher reasoning, planning, and programming complexity than other simple well-defined programming tasks. 3) Current LLMs inherently lack code validation and optimization mechanisms to ensure the correctness of synthesized programs and improve the performance of programs in terms of execution efficiency and inference accuracy in AIoT program synthesis tasks. 


To validate the above hypotheses, we conduct follow-up experiments 1) by providing necessary domain-specific knowledge to facilitate the design and implementation of corresponding algorithms to address the AIoT task and 2) by explicitly instructing the LLM to synthesize programs with clear structures via a divide-and-conquer approach.
Fig.~\ref{fig:user_intervention} illustrates the code generation process involving user intervention. We first manually retrieve relevant background knowledge (\eg, definitions of ECG data and R-peak, research papers about heartbeat detection) from the Internet and feed the information to the LLM, enabling in-context learning. Second, we instruct the LLM to learn the relevant context and comprehend the papers. Then, we ask the LLM to generate an outline of the algorithm in the paper. We further request the LLM to enrich the outline with more advanced and detailed technologies. Later, we ask the LLM to generate code snippets corresponding to each step of the algorithm. The final program is thereafter synthesized by integrating all the code snippets. Finally, we fix bugs if there are any, and execute the program to evaluate its performance with test data. We further give feedback and ask the LLM to refine the program. With several rounds of iterations, the synthesized program evolves and improves its performance in the task. 

In summary, although the synthesized programs eventually achieve reasonable performance in the tested AIoT tasks, this LLM-driven development method demands specialized domain expertise and constant manual intervention through several rounds of iterations for program optimization. 


\vspace{-9pt}
\subsection{Motivation \& Key Ideas}
\vspace{-2pt}
In this paper, we aim to develop an LLM-driven automated natural language programming system named \textit{AutoIOT} to synthesize programs for AIoT applications. \textit{AutoIOT} features three key modules: 1) \textit{Background knowledge retrieval} module that automatically collects domain knowledge from the Internet for in-context learning; 2) \textit{Automated program synthesis} module that emulates the program development lifecycle \cite{jain2015systematic} via CoT prompting. This module decomposes an AIoT task into several subtasks and generates corresponding functional code snippets; and 3) \textit{Code improvement} module that executes the synthesized program and feeds the compiler and interpreter feedback to the LLM, facilitating iterative code correction and improvement. We note that although the program synthesis process needs communication and interaction with remote LLM servers, the synthesized program can be executed locally on the client side. 
This approach fundamentally differs from existing approaches such as Penetrative AI \cite{10.1145/3638550.3641130}, and allows users to not only preserve data privacy but also improve the interpretability of synthesized programs as well as inference results. 



\vspace{-2pt}
\section{System Overview}



\vspace{-2pt}
\textit{AutoIOT} builds an intelligent agent that can automatically synthesize programs to fulfill user requirements in AIoT applications. As shown in Fig.~\ref{fig:system_overview}, \textit{AutoIOT} comprises three key modules: \textit{background knowledge retrieval} (\S~\ref{sec:background}), \textit{automated program synthesis} (\S~\ref{sec:automation}), and \textit{code improvement} (\S~\ref{sec:improvement}).

Users can specify their requirements on AIoT applications in natural language (\ding{172}). Then, the \textit{background knowledge retrieval} module identifies a set of relevant terminologies (\ding{173}) and searches over the Internet (\ding{174}). With the retrieved domain-specific knowledge, the \textit{automated program synthesis} module instructs the agent to draft an algorithm outline (\ding{175}). The agent is then requested to elaborate on each step of the algorithm and produce a detailed design (\ding{176}). Such a process decomposes a complex AIoT task into several manageable subtasks. Then, the agent is instructed to generate a code segment for each subtask (\ding{177}). Afterward, the agent is requested to integrate the codes for subtasks and synthesize the final program (\ding{178}).
Next, the \textit{code self-improvement} module executes the synthesized program and feeds the compiler and interpreter output back to the agent. The agent iteratively corrects syntax and semantics errors  (\ding{179}). With the obtained output from the synthesized program, \textit{AutoIOT} explicitly instructs the agent to explore more advanced algorithms using the web search tool, aiming to optimize the program and improve the performance of inference tasks (\ding{180}). After several iterations (\ding{175}-\ding{180}), \textit{AutoIOT} will present the final program with detailed documentation (\ding{181}). In addition, \textit{AutoIOT} provides an interface for users to offer specific algorithms or instructions for code improvement.

To enable the interaction between the LLM and the web search tool, the knowledge database, and the code executor, we leverage the LangChain \cite{langchain} framework to build an intelligent agent.
LangChain assembles various tools (abstracted as functions) and provides descriptions of the available functions (added into prompts) to the LLM. With such a prompt, the LLM performs reasoning and selects suitable tools to answer the user's query (potentially via multiple rounds of function invocations and message exchanges initiated by the LLM). This approach allows the LLM to answer queries that require context information (\eg, local weather, user's local documents), augmenting the LLM with retrieved knowledge.

\begin{figure}[t]
\vspace{-12pt}
\setlength{\abovecaptionskip}{4pt}
    \centering
    \includegraphics[width=0.45\textwidth]{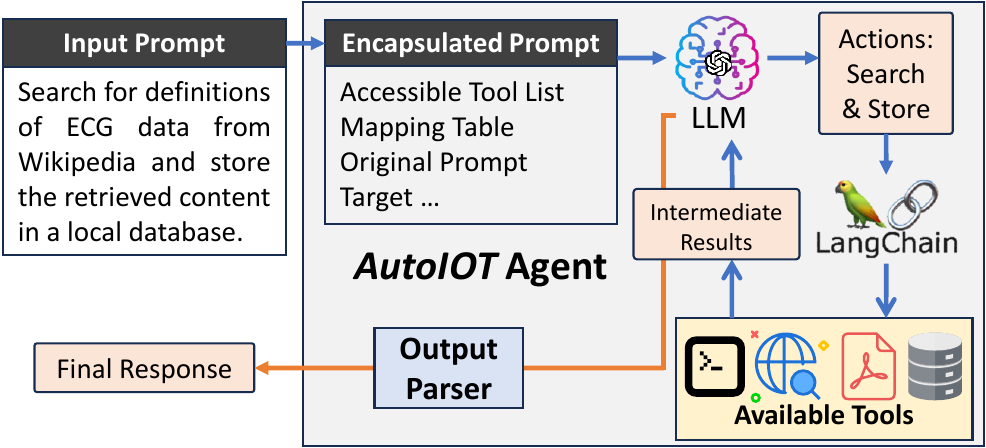}
    \caption{\textit{AutoIOT} agent answers user's prompt with LLMs and LangChain tools.}
    \label{fig:langchain}
    \vspace{-15pt}
\end{figure}

Taking step \ding{174} in Fig.~\ref{fig:system_overview} as an example, Fig.~\ref{fig:langchain} shows how \textit{AutoIOT} agent works with LangChain tools and the LLM to answer the query about terminology searching via network.
Given an input prompt, \textit{AutoIOT}
encapsulates it with additional information (\eg, a list of available tools) and sends it to the LLM. Then, the LLM performs a sequence of actions possibly leveraging the available tools in the list, and generates an output. 
The output will then be passed to a parser to generate the final response. We note that the above processes involved in step \ding{174} as well as other steps (\ding{172}-\ding{181} in Fig.~\ref{fig:system_overview}) are all automatically orchestrated by the \textit{AutoIOT} agent.

\textbf{Usage Scenario.} Suppose a user wants to develop a heartbeat detection application, she can interact with \textit{AutoIOT} with natural language, which describes her requirements for the application. Then, \textit{AutoIOT} will automatically synthesize a corresponding program and documentation for the user. Following the instructions in the documentation, she can deploy and execute the program on a target device, which contains the patient's ECG data. The program will then generate the final heartbeat detection results for her.

\vspace{-8pt}
\section{System Design}

\vspace{-2pt}
In this section, we select heartbeat detection as an example to illustrate how \textit{AutoIOT} works. 
%
\vspace{-5pt}
\subsection{User Interface}
\vspace{-2pt}
Users can describe an AIoT task in natural language as input to \textit{AutoIOT}. To help the LLM interpret the intention and desired outcome (\ie, synthesized program and inference results), we design a prompt template for the user to describe the problem, since LLMs can comprehend and process well-structured instructions more efficiently \cite{dunn2022structured}. As shown in Fig.~\ref{fig:system_overview}, the user problem includes four parts: target, remarks, and program input and output specifications. The target part describes the user's objective and task in natural language, \eg, "Given the MIT-BIH Arrhythmia Database, please load all the ECG records and detect all the R-peaks. Then, evaluate the detection results and output the detection accuracy for each record." The remarks provide additional information, \eg, "I only downloaded a zip file from the dataset's official website". The program input and output specifications clarify the I/O format of the synthesized program. The path to the dataset is required during the \textit{code improvement} process for program execution and optimization. 


\vspace{-8pt}
\subsection{Background Knowledge Retrieval}
\label{sec:background}
\vspace{-2pt}
LLMs are typically trained on extensive and pre-collected corpus datasets that include a wide range of general common knowledge over the Internet. These training datasets, however, may not include domain-specific knowledge or the latest advances in research literature.
In the rapidly evolving AIoT field, with new technologies and algorithms constantly emerging, the knowledge gap is particularly pronounced. To fill this gap, we develop the \textit{background knowledge retrieval} module to automatically identify and fetch necessary information online so as to enable in-context learning for LLMs augmented with up-to-date domain knowledge.

\begin{figure*}[t]
\vspace{-8pt}
\setlength{\abovecaptionskip}{-4pt}
\subfigtopskip=-4pt
\subfigcapskip=-4pt
    \centering
    \subfigure[Terminology determination]{
        \label{fig:terminology_determination}
        \includegraphics[width=0.225\textwidth]{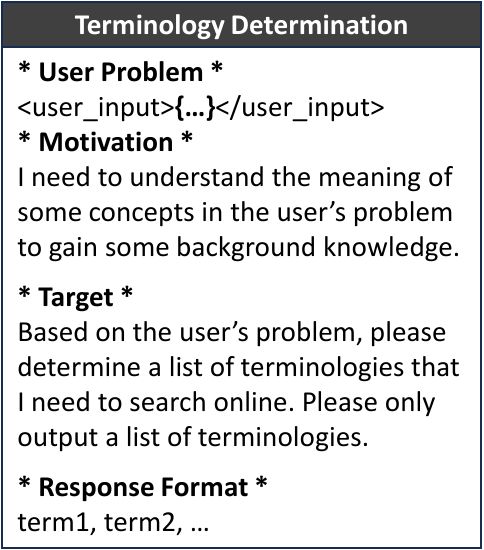}
    }
    \centering
    \subfigure[Terminology searching]{
        \label{fig:terminology_searching}
        \includegraphics[width=0.225\textwidth]{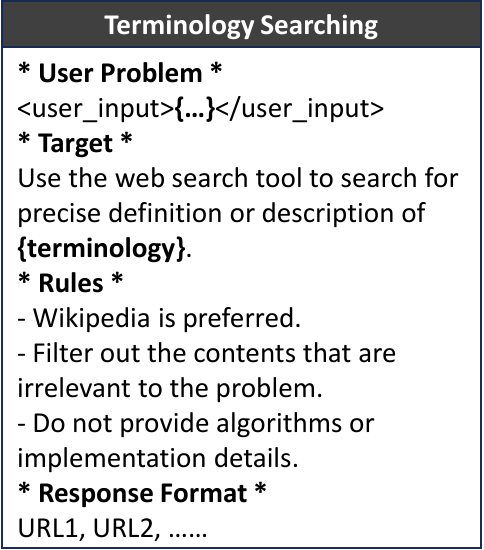}
    }
    \centering
    \subfigure[Algorithm outline generation]{
        \label{fig:algorithm_outline}
        \includegraphics[width=0.225\textwidth]{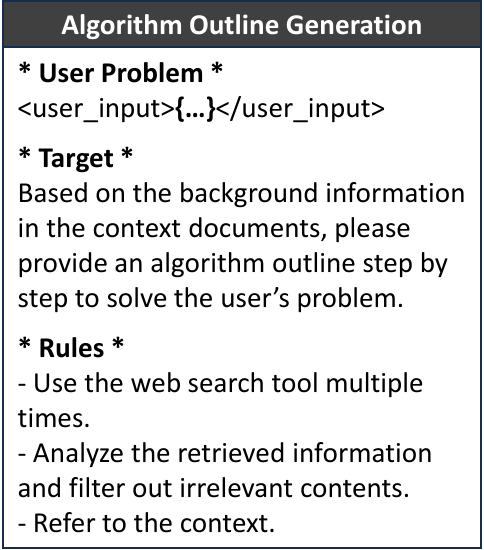}
    }
    \centering
    \subfigure[Detailed design generation]{
        \label{fig:algorithm_detail}
        \includegraphics[width=0.225\textwidth]{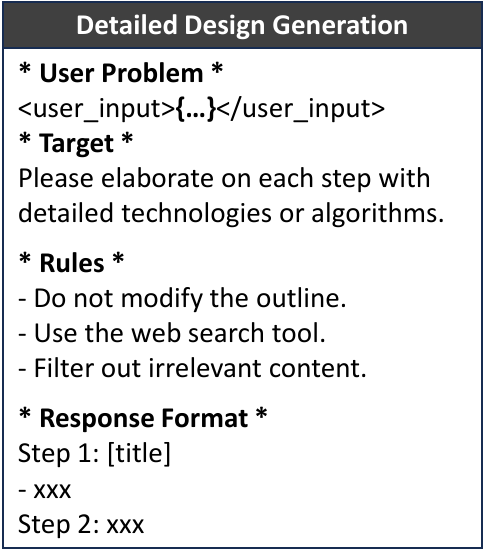}
    }
    \caption{The prompt template for (a) terminology determination, (b) terminology searching, (c) algorithm outline generation, and (d) detailed design generation.}
    \label{fig:prompt1}
    \vspace{-18pt}
\end{figure*}

\noindent\textbf{Terminology Determination.} The \textit{background knowledge retrieval} module first instructs the agent to identify some relevant key terminologies given the user problem, with the prompt shown in Fig.~\ref{fig:terminology_determination}. For example, given the user problem in Fig.~\ref{fig:system_overview}, the terminologies generated by the LLM are "MIT-BIH Arrhythmia Database", "ECG data", and "R-peaks". Next, to obtain the relevant knowledge, \textit{AutoIOT} instructs (Fig.~\ref{fig:terminology_searching}) the LLM to actively search for the definitions and descriptions of these terminologies from public websites, such as Wikipedia and GitHub, with the web search tool. Additionally, for the terminologies with multiple interpretations, \textit{AutoIOT} requests the LLM to filter out the irrelevant content and focus on those pertinent to the user problem.

\noindent\textbf{Context Database Construction.} After retrieving necessary information from relevant websites, \textit{AutoIOT} uses OpenAI's text embedding \cite{neelakantan2022text} to convert the HTML documents into vector representations, containing semantic meanings comprehensible to the LLM. These representations are then used to build a local vector database with Faiss \cite{douze2024faiss}, serving as a contextual knowledge base for the LLM. During the inference process, the LLM retrieves relevant content from the database with a high degree of similarity to the user problem in the vector space.
This approach ensures that the LLM understands the user problem and objective with necessary context information and domain knowledge.


\noindent\textbf{Remarks.} The \textit{background knowledge retrieval} module is user-friendly, as it is operated automatically by our \textit{AutoIOT} agent without any user intervention. In addition, we provide an interface for users to explicitly complement necessary background information as well as highly specialized domain knowledge (\eg, research papers, detailed algorithm descriptions) to enrich the context database. 
\vspace{-10pt}
\subsection{Automated Program Synthesis}
\label{sec:automation}
\vspace{-2pt}
As observed in our preliminary experiments (\S~\ref{sec:preliminary}), if we directly instruct LLMs to generate programs for AIoT applications, multiple subtasks should be undertaken manually by the user. Specifically, the user needs to decompose the AIoT task into several subtasks and request the LLM to generate a solution for each subtask. After integrating the solutions, the user has to manually debug, execute, and improve the program. Although the synthesized program eventually meets the user's requirement, the program synthesis necessitates frequent user intervention and active involvement throughout the process, which is cumbersome and time-consuming. To address this problem, we develop the \textit{automated program synthesis} module, aiming to automate the programming procedure, reduce the involved workload, and improve the development efficiency and user experience. 
In particular, the \textit{automated program synthesis} module uses Chain-of-Thought (CoT) prompts to guide LLMs through step-by-step reasoning processes, thereby 
enhancing their capability of tackling complex AIoT problems by mimicking human-like divide-and-conquer reasoning processes.

\noindent\textbf{CoT 1: Algorithm Outline Generation.} \textit{AutoIOT} first prompts the LLM to analyze the user problem and design a preliminary algorithm outline. As illustrated in Fig.~\ref{fig:algorithm_outline}, the prompt for algorithm outline generation consists of three parts: 1) The "User Problem" is reiterated at the beginning to ensure continuity and coherence in the LLM's responses since the LLM may forget the previous context \cite{wang2023recursively}. 
2) The "Target" specifies our request, \ie, algorithm outline generation. 3) The "Rules" add detailed instructions for quality assurance. 
For example, \textit{AutoIOT} explicitly requests the LLM to actively search for advanced AIoT algorithms using the web search tool throughout the process. 
\textit{AutoIOT} also asks the LLM to filter out irrelevant information. With such well-structured prompts, the LLM can generate an algorithm outline according to the problem specification.

\noindent\textbf{CoT 2: Detailed Design Generation.} Given the generated algorithm outline, the LLM is then tasked with further elaborating on each step in the outline with more detailed technologies and algorithms to refine the approach comprehensively. The prompt for this stage includes "User Problem", "Target", and "Rules". In addition, it also includes a new requirement - "Response Format" to specify the expected format of the LLM's output, as illustrated in Fig.~\ref{fig:algorithm_detail}. 
Given such a prompt, the LLM can generate detailed steps and specific actions in each step to achieve the overall objective and solve the user problem. In this stage, essentially \textit{AutoIOT} guides the LLM to decompose the AIoT task into multiple subtasks, facilitating a divide-and-conquer strategy to synthesize the corresponding program in the next stage.


\begin{figure*}[t]
\vspace{-8pt}
\setlength{\abovecaptionskip}{-4pt}
\subfigtopskip=-4pt
\subfigcapskip=-4pt
    \centering
    \subfigure[Modularized code generation]{
        \label{fig:modularized_code}
        \includegraphics[width=0.225\textwidth]{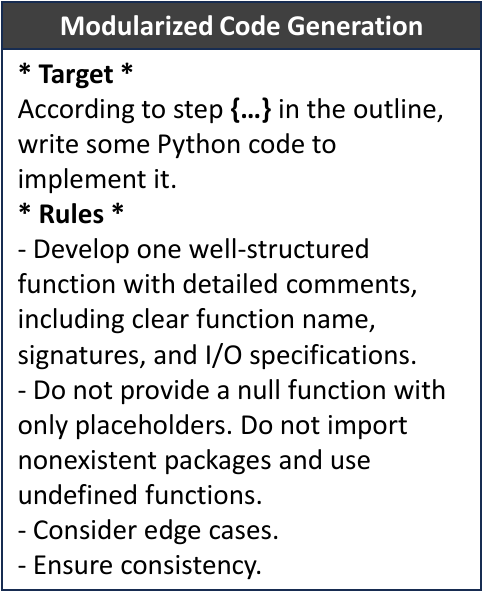}
    }
    \centering
    \subfigure[Modularized code integration]{
        \label{fig:code_integration}
        \includegraphics[width=0.225\textwidth]{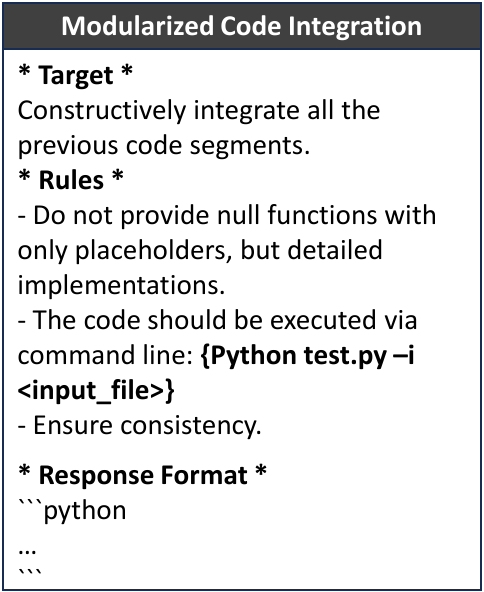}
    }
    \centering
    \subfigure[Code debugging]{
        \label{fig:code_debugging}
        \includegraphics[width=0.225\textwidth]{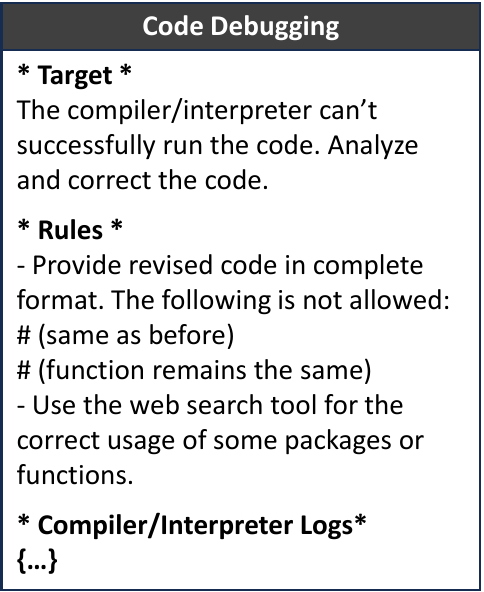}
    }
    \centering
    \subfigure[Algorithm modification]{
        \label{fig:code_improvement}
        \includegraphics[width=0.225\textwidth]{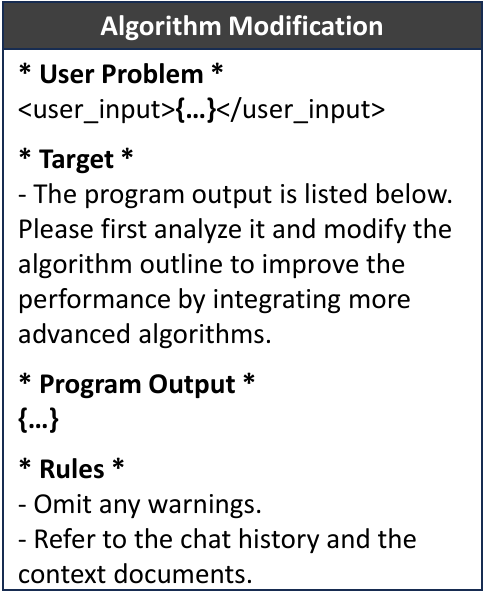}
    }
    \caption{The prompt template for (a) modularized code generation, (b) modularized code integration, (c) code debugging, and (d) code improvement via algorithm modification}
    \label{fig:prompt2}
    \vspace{-15pt}
\end{figure*}

\noindent\textbf{CoT 3: Modularized Code Generation.} 
Given a set of subtasks generated at the previous stage, \textit{AutoIOT} instructs the LLM to generate one function for each module with a clear function name, signature, and input/output specification.
The modules can then be developed independently, ensuring specific functionalities are effectively implemented according to the algorithm descriptions generated above. This divide-and-conquer approach is proven effective in synthesizing complex programs
by fully exploiting the modularized code generation capabilities \cite{austin2021program} of LLMs. 

In our initial trials, we observed that LLMs sometimes generate null functions with placeholders, invoke undefined functions, or import nonexistent packages. To tackle this problem, \textit{AutoIOT} adds more stringent rules and requirements to explicitly ask the LLM to avoid generating null functions and verify the availability of imported packages and invoked functions by involving the web search tool. 
With the prompts shown in Fig.~\ref{fig:modularized_code}, the LLM can generate cohesive code segments with detailed comments for each module, facilitating the module integration in the next stage. 

\noindent\textbf{CoT 4: Modularized Code Integration.} 
Given the generated code segments, \textit{AutoIOT} prompts the LLM to constructively integrate all modularized code segments and create a cohesive and comprehensive program.
Since the code generated for different modules may have disparate input/output variable names, \textit{AutoIOT} first prompts the LLM to ensure the consistency among all the modules and synthesize the final program without null functions as illustrated in Fig.~\ref{fig:code_integration}.

For the convenience of code execution, debugging, and optimization, \textit{AutoIOT} asks the LLM to add a main function so that the program can be directly executed from the command console (\eg, \verb|python3 test.py -i <input_file>|). Moreover, \textit{AutoIOT} also asks the LLM to generate user documentation in Markdown format, specifying how to properly install, execute, and troubleshoot the program for end users.

\noindent\textbf{Remarks.} The \textit{automated program synthesis} module facilitates a seamless transformation from natural language to a readily executable program with CoT prompts. \textit{AutoIOT} is regarded as an experienced developer, adept at decomposing complex AIoT tasks into multiple modules, generating modularized code, and organically integrating them. The program synthesis process can be fully automated by the agent.

\vspace{-5pt}
\subsection{Code Improvement}
\label{sec:improvement}
\vspace{-2pt}
In \S~\ref{sec:preliminary}, we found that the LLM can evaluate and improve the code with heavy user intervention. 
To alleviate the user's workload involved in debugging and code optimization, we develop the \textit{code improvement} module.


\noindent\textbf{Automated Debugging.} Upon obtaining the final program after integration, \textit{AutoIOT} constructs a code executor to run the generated code within a virtue environment (\eg, a sandbox), ensuring safe and controlled code execution. 
The code executor loads the sensor dataset from the user's local device for program execution and exports the compiler or interpreter output to the LLM.
If the program encounters execution issues (\eg, syntax or I/O errors), \textit{AutoIOT} embeds the logs from the compiler into a prompt (Fig.~\ref{fig:code_debugging}) and instructs the LLM to debug the code for several rounds of interactions 
until the generated code can be executed successfully. 

\noindent\textbf{Code Optimization via Algorithm Modification.} 
To achieve better performance, 
\textit{AutoIOT} progressively refines the synthesized program via several iterations.
In particular, \textit{AutoIOT} processes the test dataset with the first version of the integrated program. 
Then, \textit{AutoIOT} prompts (Fig.~\ref{fig:code_improvement}) the LLM with the context information (\eg, algorithm outline, chat history) of generating the first program as well as the program output, and asks the LLM to improve the performance by integrating more advanced algorithms. Specifically, \textit{AutoIOT} uses the web search tool to search for solutions that can achieve higher accuracies referring to academic papers and relevant webpages.
This initiates a new recursive cycle of program synthesis, starting from refining the algorithm outline accordingly, enriching the outline with the retrieved algorithms, generating modularized code for the updated design, combing the modularized code, debugging, and improving code quality. 
This code optimization cycle is not a one-time process but is repeated multiple times. Empirically,
\textit{AutoIOT} takes five iterations to progressively generate five different programs, striking a balance between thoroughness and efficiency. Finally, \textit{AutoIOT} requests the LLM to analyze the execution results of all the programs and select the one that achieves the best performance as the final program. 

\begin{figure*}[t]
\vspace{-12pt}
\setlength{\abovecaptionskip}{4pt}
    \centering
    \includegraphics[width=\textwidth]{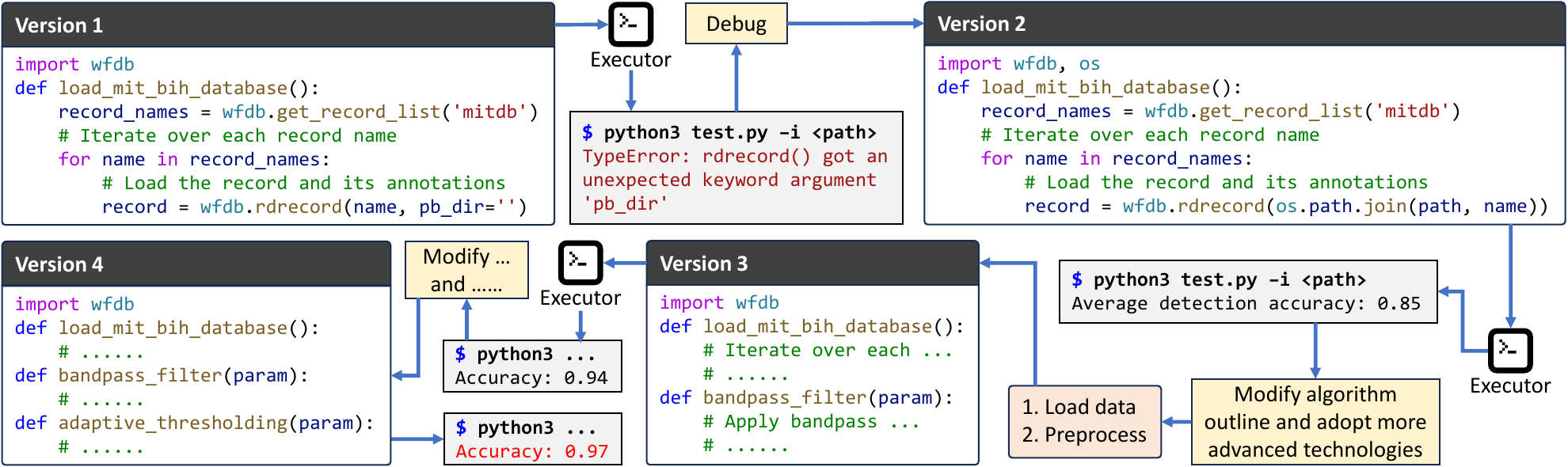}
    \caption{An example of code improvement via iterations (details omitted).}
    \label{fig:code_improvement_sample}
    \vspace{-18pt}
\end{figure*}

Fig.~\ref{fig:code_improvement_sample} shows a specific example of how the synthesized program evolves over multiple iterations.
Version 1 is generated with a bug. By providing the error message to the LLM, \textit{AutoIOT} can automatically fix the bug and generate Version 2.
\textit{AutoIOT} further instructs the LLM to modify the algorithm and adopt advanced technologies to generate new versions that iteratively achieve higher accuracies.

\noindent\textbf{Supporting User-in-the-Loop Optimization.} After each iteration of code optimization, \textit{AutoIOT} also provides an interface for the user to \textit{optionally} provide instructions that can help the LLM improve the synthesized program. For example, when the user finds that the LLM fails to recall relevant information from the retrieved contents, the user can prompt  \textit{AutoIOT} to refer to a specified algorithm or provide a recent academic paper. This enables a user-in-the-loop optimization that requires minimal user intervention and promotes code optimization iteratively.



\noindent\textbf{Remarks.} The \textit{code improvement} module automates code debugging and optimization by leveraging the LLM's proficiency in debugging and refining code based on the compiler and interpreter feedback \cite{chen2023teaching, madaan2024self}. 
This automation not only releases the manual burden but also heralds a new era where AIoT applications can evolve iteratively and autonomously with minimum user intervention.




\vspace{-8pt}
\section{Experiment Setup}
\vspace{-3pt}
\subsection{Implementation}
\vspace{-2pt}
We implement \textit{AutoIOT} with GPT-4 \cite{achiam2023gpt} based on LangChain \cite{langchain}, which provides various tools (\eg, web search engine, vector database, \etc) for LLMs to collect relevant information. 
We select Tavily \cite{tavilyai} as the web search tool to search for relevant information. 
It uses OpenAI's text embedding model \cite{neelakantan2022text} to convert the retrieved webpages into vector representations. 
\textit{AutoIOT} then uses Faiss \cite{douze2024faiss} for efficient similarity search of vector representations.  
The code executor controlled by \textit{AutoIOT} is deployed on a Linux Ubuntu workstation equipped with an NVIDIA RTX 4090 GPU.

\vspace{-5pt}
\subsection{AIoT Applications \& Datasets}
\vspace{-2pt}
We select four representative AIoT applications from the domains of healthcare and human activity recognition (HAR). Unlike other program synthesis tasks, these AIoT tasks require domain-specific knowledge and highly specialized algorithms in signal processing and machine learning.

\noindent\textbf{Heartbeat Detection.} R-peak detection in electrocardiogram (ECG) data is a crucial task in cardiac signal processing, serving as a foundational step for heart rate variability studies, and arrhythmia detection \cite{wu2021ecg}. We use MIT-BIH Arrhythmia Database \cite{moody2001impact} and five representative baseline algorithms, including Hamilton \cite{hamilton2002open}, Christov \cite{christov2004real}, Engzee \cite{engelse1979single}, Pan-Tompkins \cite{pan1985real}, and SWT \cite{kalidas2017real}.

\noindent\textbf{IMU-based Human Activity Recognition.} Inertial measurement unit (IMU)-based HAR enables continuous identification of a wide range of daily activities (\eg, sitting, walking) by capturing and analyzing motion characteristics from the IMU data \cite{gao2023exploring, xu2021limu}. For the baselines, we select five open-source GitHub repositories: LSTM-RNN \cite{venelin2017}, 1D-CNN \cite{Akshay2019}, Conv-LSTM \cite{jimmy2020}, BiLSTM \cite{Tomasz2020}, and NN \cite{Athanasiou2022}. We compare their performance with \textit{AutoIOT} on the WISDM dataset \cite{kwapisz2011activity}.

\noindent\textbf{mmWave-based Human Activity Recognition.} mmWave can capture fine-grained human gestures with high resolution \cite{xue2022m4esh, cui2024talk2radar, cui2023mmripple}. We select the XRF55 dataset \cite{wang2024xrf55} with the models proposed in the paper as the baselines, including ResNet-18, 34, 50, 101, and 152.
This task is more challenging because: 1) The XRF55 dataset was recently published on websites only a few months ago, which means LLMs have not yet seen knowledge about this dataset; 2) The mmWave data has high dimensionality, necessitating the use of more sophisticated models with optimized configurations. 


\noindent\textbf{Multimodal Human Activity Recognition.} By leveraging different sensors to capture complementary information, HAR systems can achieve higher robustness and versatility. We select the Harmony dataset \cite{ouyang2023harmony} containing three sensor modalities: audio, depth image, and radar. The baseline system consists of three encoders, each designed to extract unique features from the respective modalities, followed by feature concatenation and a classifier model. This task is also challenging for \textit{AutoIOT} as it involves the fusion of data from different modalities with cross-modal interaction.

\vspace{-8pt}
\section{Evaluation}
\vspace{-3pt}
\subsection{Metrics}
\vspace{-2pt}

\begin{figure*}[t]
\vspace{-8pt}
\setlength{\abovecaptionskip}{-4pt}
\subfigtopskip=-6pt
\subfigcapskip=-6pt
    \centering
    \subfigure[Heartbeat detection]{
        \label{fig:overall_acc_mae}
        \includegraphics[width=0.225\textwidth]{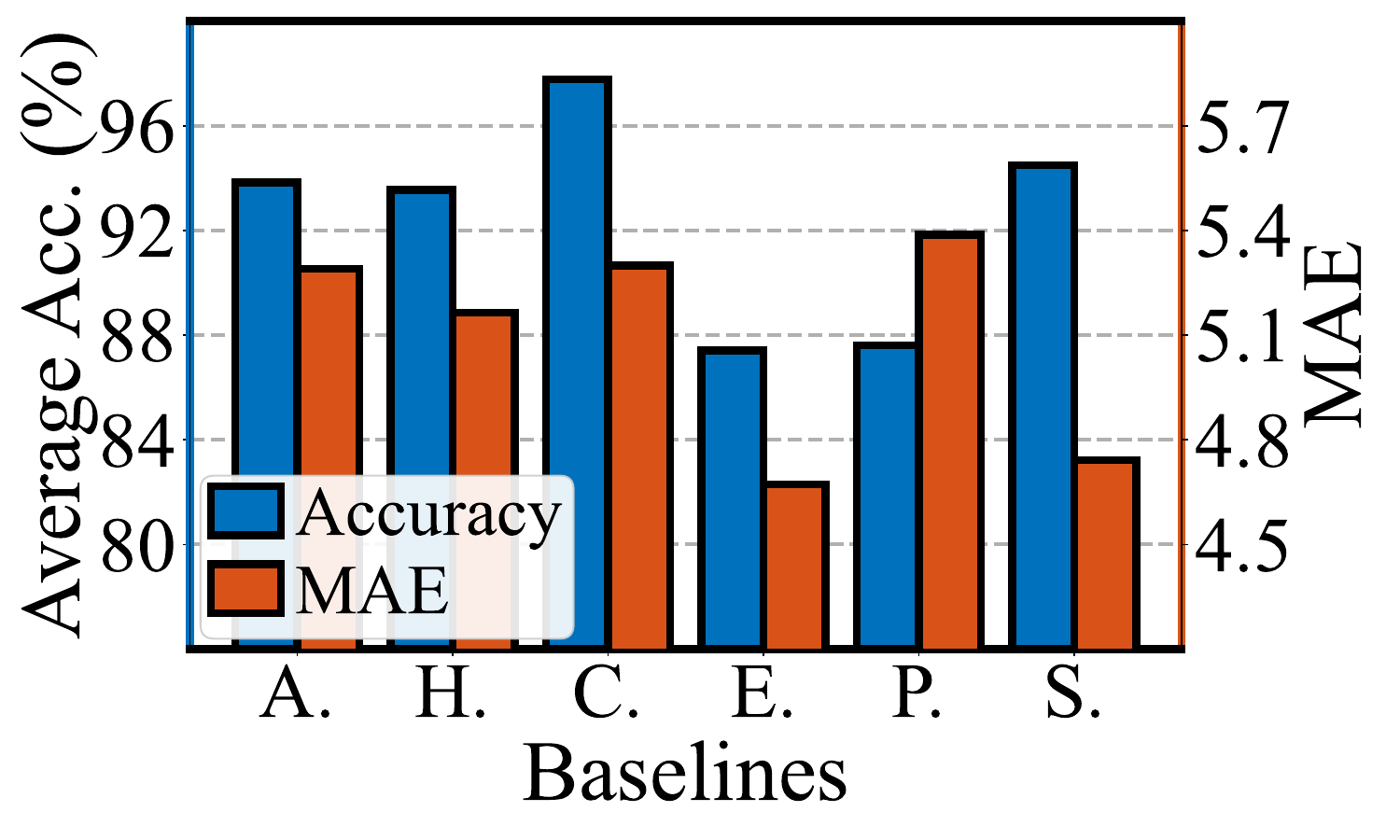}
    }
    \centering
    \subfigure[IMU \& mmWave-based HAR]{
        \label{fig:overall_acc_HAR}
        \includegraphics[width=0.225\textwidth]{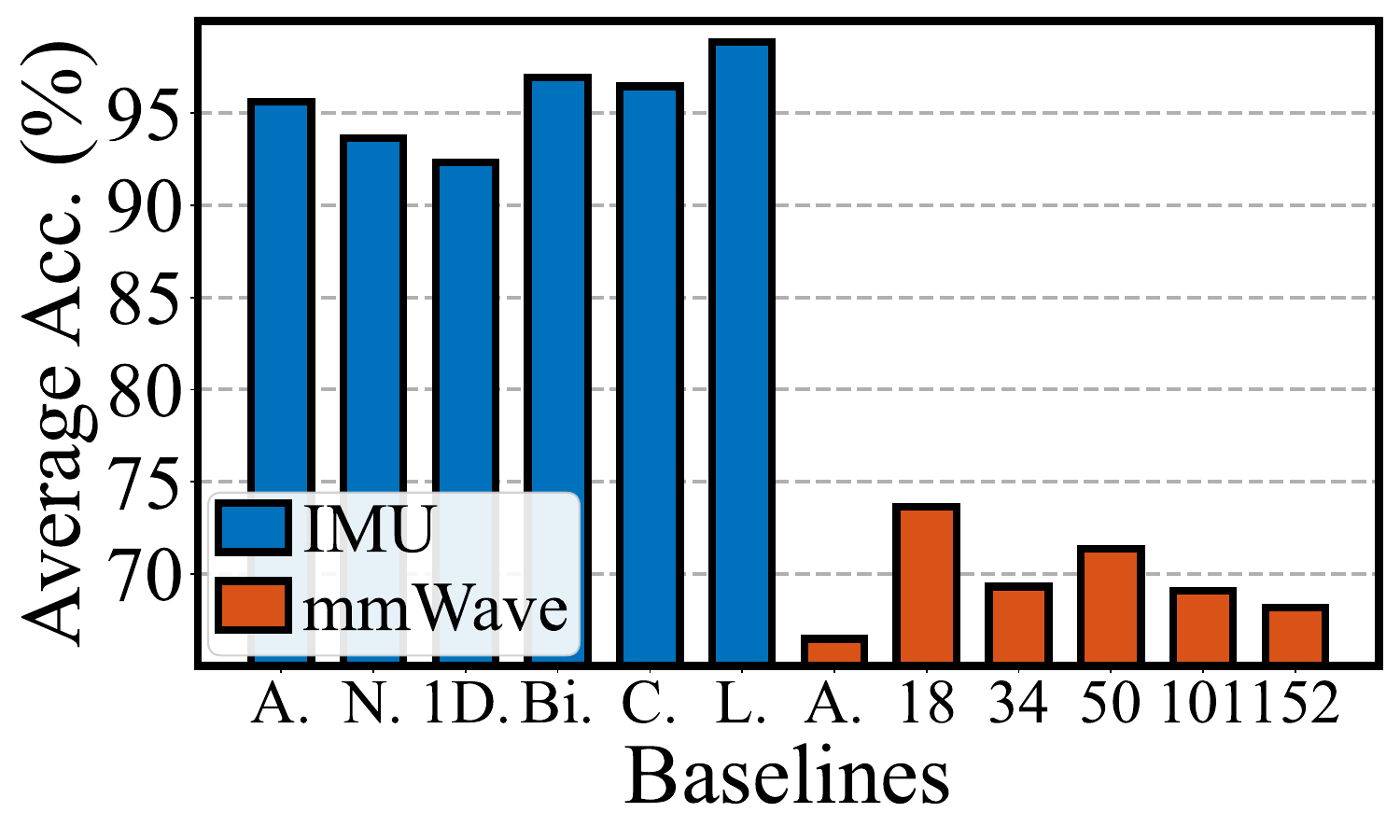}
    }
    \centering
    \subfigure[Multimodal HAR]{
        \label{fig:overall_multimodal}
        \includegraphics[width=0.225\textwidth]{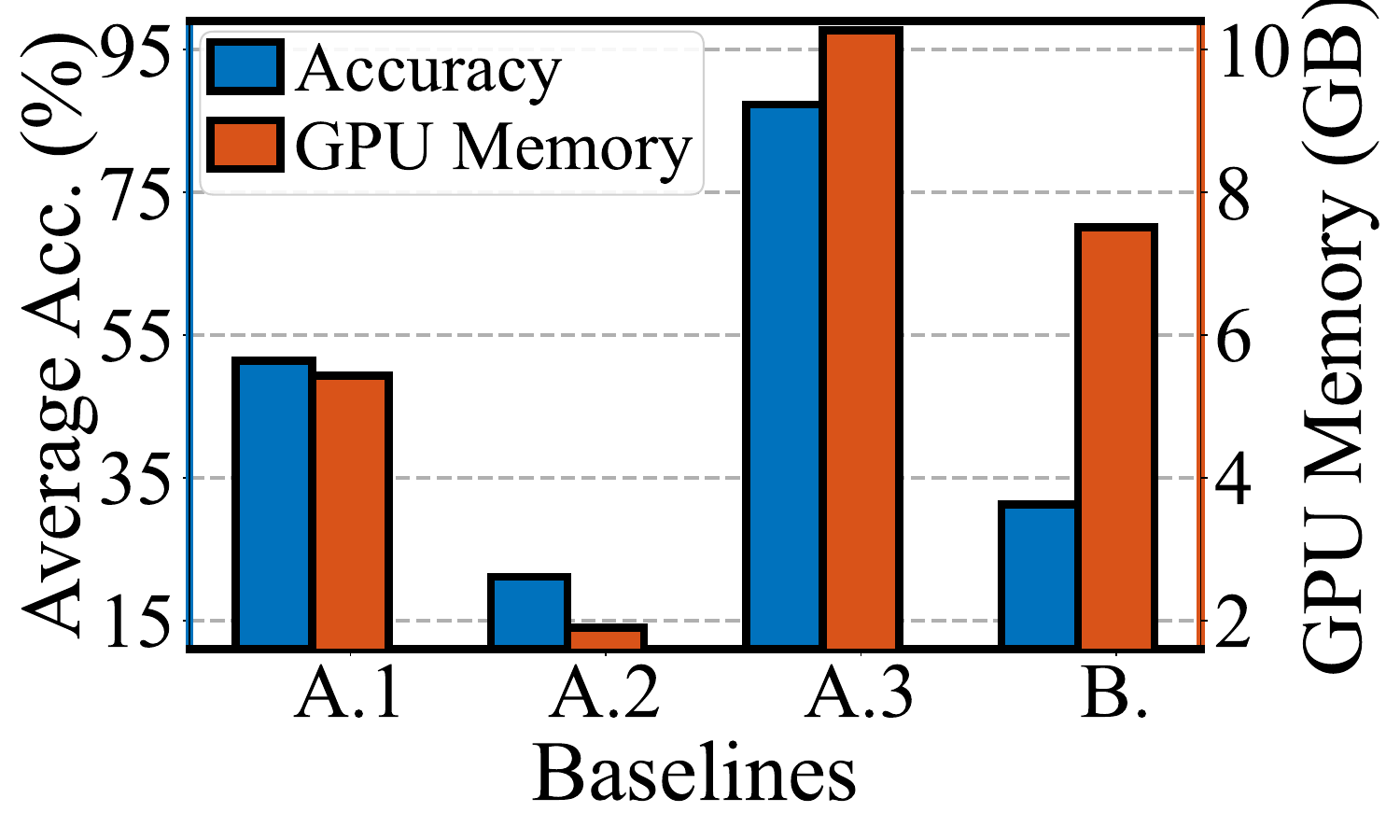}
    }
    \centering
    \subfigure[Inference time per sample]{
        \label{fig:memory_time}
        \includegraphics[width=0.225\textwidth]{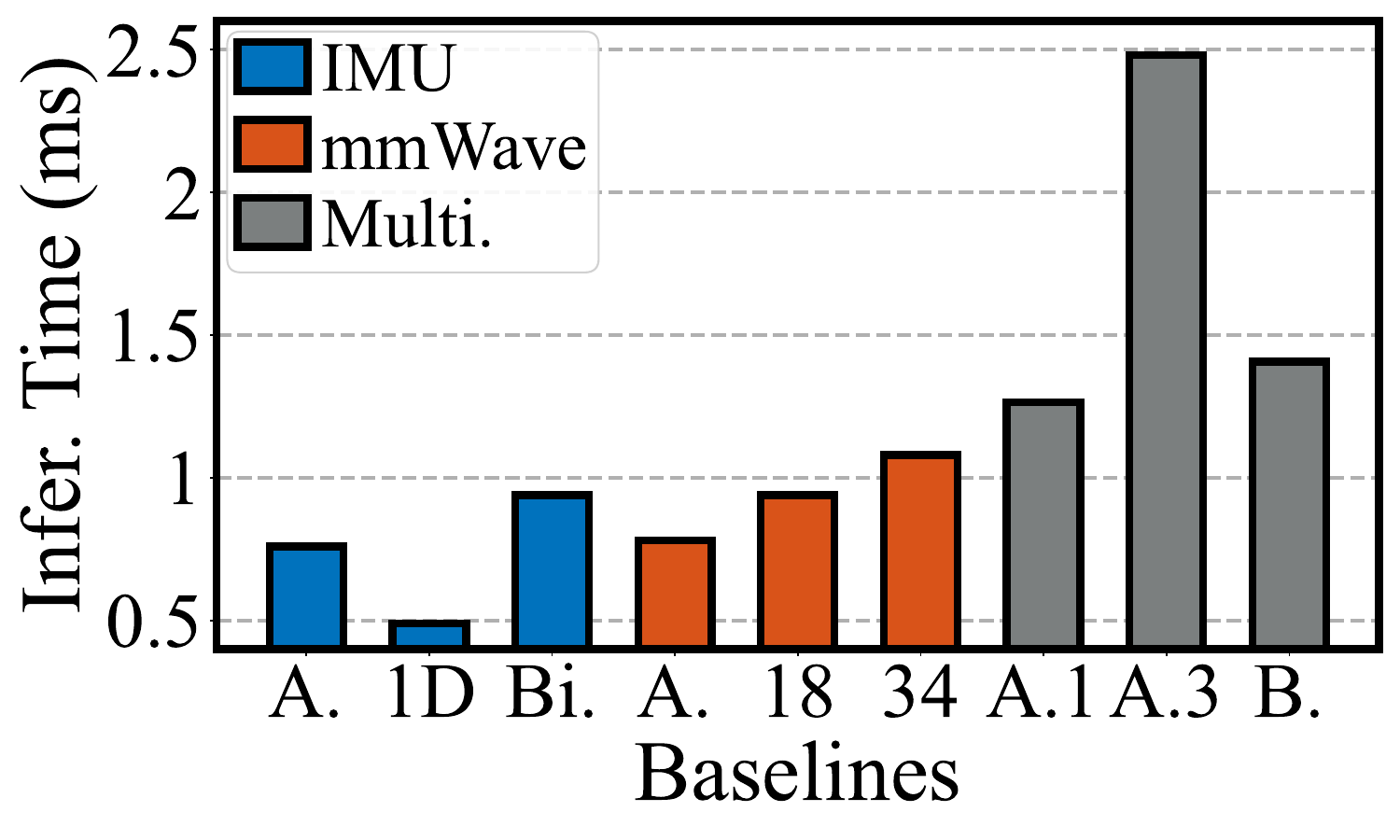}
    }
    \caption{The overall performance of the four IoT applications. In (a), A. for \textit{AutoIOT}, H. for Hamiltion, C. for Christov, E. for Engzee, P. for Pan-Tompkins, and S. for SWT. In (b), N. for NN, 1D for 1D-CNN, Bi. for BiLSTM, C. for Conv-LSTM, L. for LSTM-RNN, and $\boldsymbol{n}$ for ResNet-$\boldsymbol{n}$. In (c) \& (d), A.1, A.2, and A.3 for three different \textit{AutoIOT}-generated programs; B. for the baseline in the multimodal HAR application.}
    \vspace{-18pt}
\end{figure*}

We adopt the following evaluation metrics: 1) \textit{Task accuracy}: we repeat the experiment 10 times and calculate the average task accuracy. In heartbeat detection, we consider task accuracy as the percentage of correctly identified peaks within a predefined tolerance window compared to the ground truth. In HAR, we consider classification accuracy. 2) \textit{MAE}: We use medium absolute error (MAE) to measure the discrepancy in beat positions between the predicted R-peaks and the ground truth. 3) \textit{Communication cost}: we use \textit{psutil} \cite{psutil} to monitor the network traffic.
4) \textit{Wall-clock execution time}: we record the total time consumed from the moment the user inputs the problem to the generation of the final inference results for all the sensor data. 
5) \textit{Memory consumption}: we record the GPU memory consumption during code execution if AI models are adopted. 6) \textit{Inference time per sample}: we compute the inference time per data sample if AI models are used.

\vspace{-10pt}
\subsection{Performance against Baselines}
\vspace{-2pt}
\subsubsection{Average accuracy \& MAE}
Fig.~\ref{fig:overall_acc_mae} shows the heartbeat detection accuracy with MAE of \textit{AutoIOT} (denoted as A.) and baselines. First of all, \textit{AutoIOT} can synthesize a program automatically to achieve comparable performance with baselines in the heartbeat detection task. More surprisingly, the automatically synthesized program can even beat some of the baselines! For example, the synthesized program achieves higher detection accuracy than Pan-Tompkins (P.) and Engzee (E.). Moreover, it yields a lower error rate than Christov (C.) and Pan-Tompkins (P.).
To investigate the reasons behind this, we examine and analyze the synthesized program. 

\textbf{We learned}: 1) Armed with the web search tool, the synthesized program implemented a few basic as well as sophisticated signal processing methods, including bandpass filtering and stationary wavelet transformation in preprocessing, and adaptive thresholding in postprocessing. 
Some selected algorithms are well-known and widely adopted, while others are less likely to be chosen, even by experienced programmers with domain expertise. 
Unlike narrowly focused, well-defined simple programming tasks, AIoT tasks typically require systematic integration of multiple algorithms and components to achieve optimal system performance, which creates opportunities for \textit{AutoIOT} to explore more possibilities in automatically synthesizing optimized programs that could outperform not all but some representative baselines.  

2) Given a single performance objective, we notice that the LLM carries out extensive optimization, sometimes at the cost of other equally important metrics. 
For example, when instructed to improve the detection accuracy, the synthesized program sets a larger tolerance window, which increases the chance of correctly detecting heartbeats (true positives) at the cost of increased false positives.
Considering AIoT applications' complexity and multiple competing or even contradicting objectives, user requirement specification needs to be as complete and comprehensive as possible, which necessitates domain expertise and system development experience.  

3) We found that the webpages returned by the web search tools are often about general algorithms due to their popularity and higher page rankings. Such popular algorithms, however, may not perform the best in domain-specific tasks. 
With minimum user intervention by providing specialized algorithms, \textit{AutoIOT} can synthesize programs accordingly and achieve comparable performance to the baselines.

Fig.~\ref{fig:overall_acc_HAR} shows the classification accuracy in two HAR tasks. We observe that \textit{AutoIOT} outperforms NN and 1D-CNN while underperforms BiLSTM, Conv-LSTM and LSTM-RNN. The main reasons are twofold: 1) HAR tasks require both signal processing and machine learning algorithms, increasing the programming complexity to some extent; 2) Training neural networks requires fine-tuning of a vast array of hyper-parameters (\eg, network architecture configurations, epoch number, learning rate, optimizer, and loss function). This significantly amplifies the instability of the generated code and calls for careful fine-tuning to achieve the best performance in practice. As a result, \textit{AutoIOT} surpasses those baselines adopting simple model architectures (NN and one-dimensional CNN) but falls short against baselines using sophisticated architectures (BiLSTM and Conv-LSTM) with highly optimized hyper-parameters. Although during code improvement, some synthesized programs define a set of configurations and adopt a searching strategy to obtain optimal hyper-parameters, the performance still remains slightly lower than some baselines. This is because the determination of the optimal configurations for machine learning models is typically a trial-and-error process, requiring substantial human effort.
Fortunately, we observe that if the user provides a potential search space in advance, the LLM can design a search algorithm to try different hyper-parameter configurations and select the one with the best performance.

\begin{figure*}[t]
\vspace{-12pt}
    \centering
    \begin{minipage}[t]{0.49\textwidth}
        \setlength{\abovecaptionskip}{-4pt}
        \subfigtopskip=-6pt
        \subfigcapskip=-6pt
        \centering
        \subfigure[Single level intervention]{
            \label{fig:intervention_single}
            \includegraphics[width=0.45\textwidth]{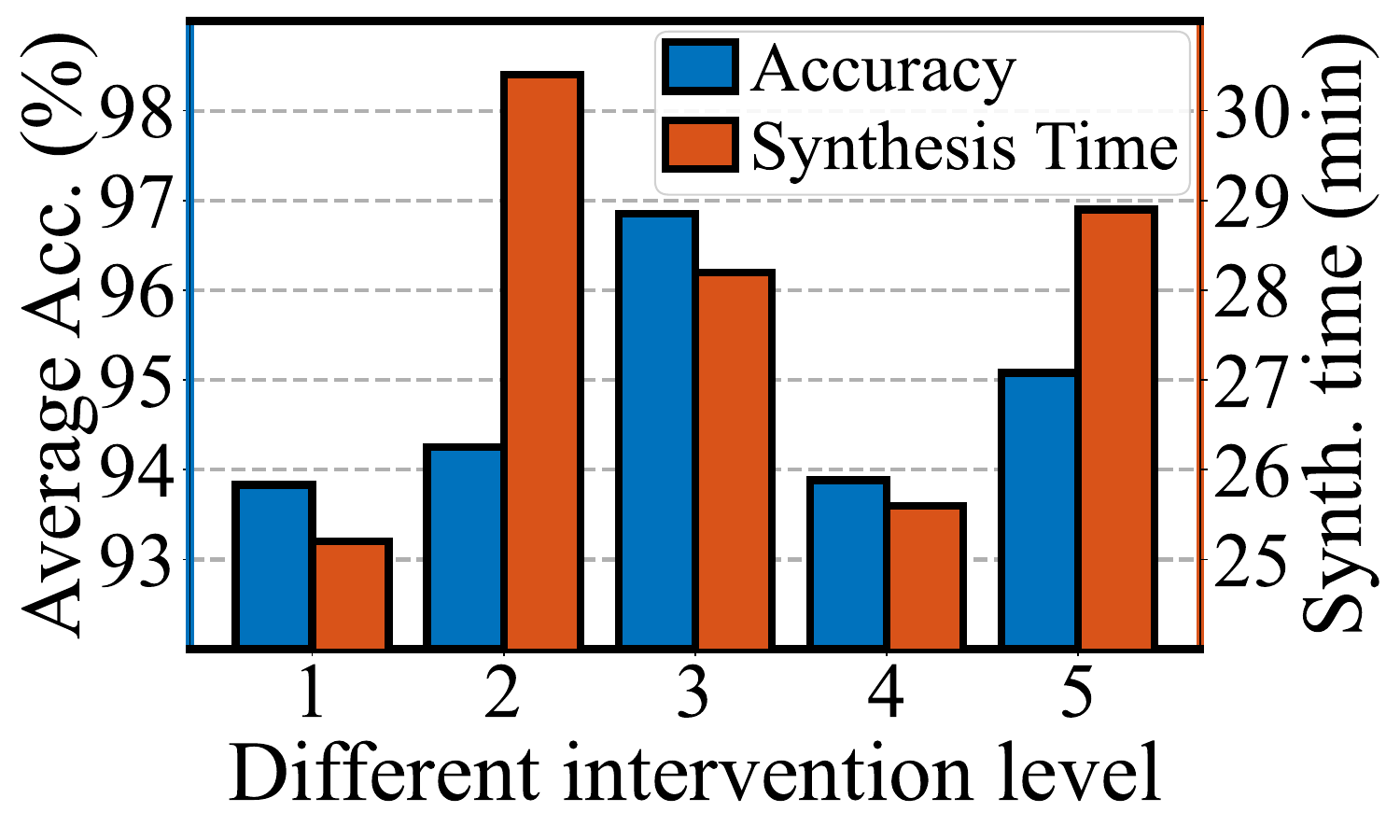}
        }
        \centering
        \subfigure[Combined intervention]{
            \label{fig:intervention_combined}
            \includegraphics[width=0.45\textwidth]{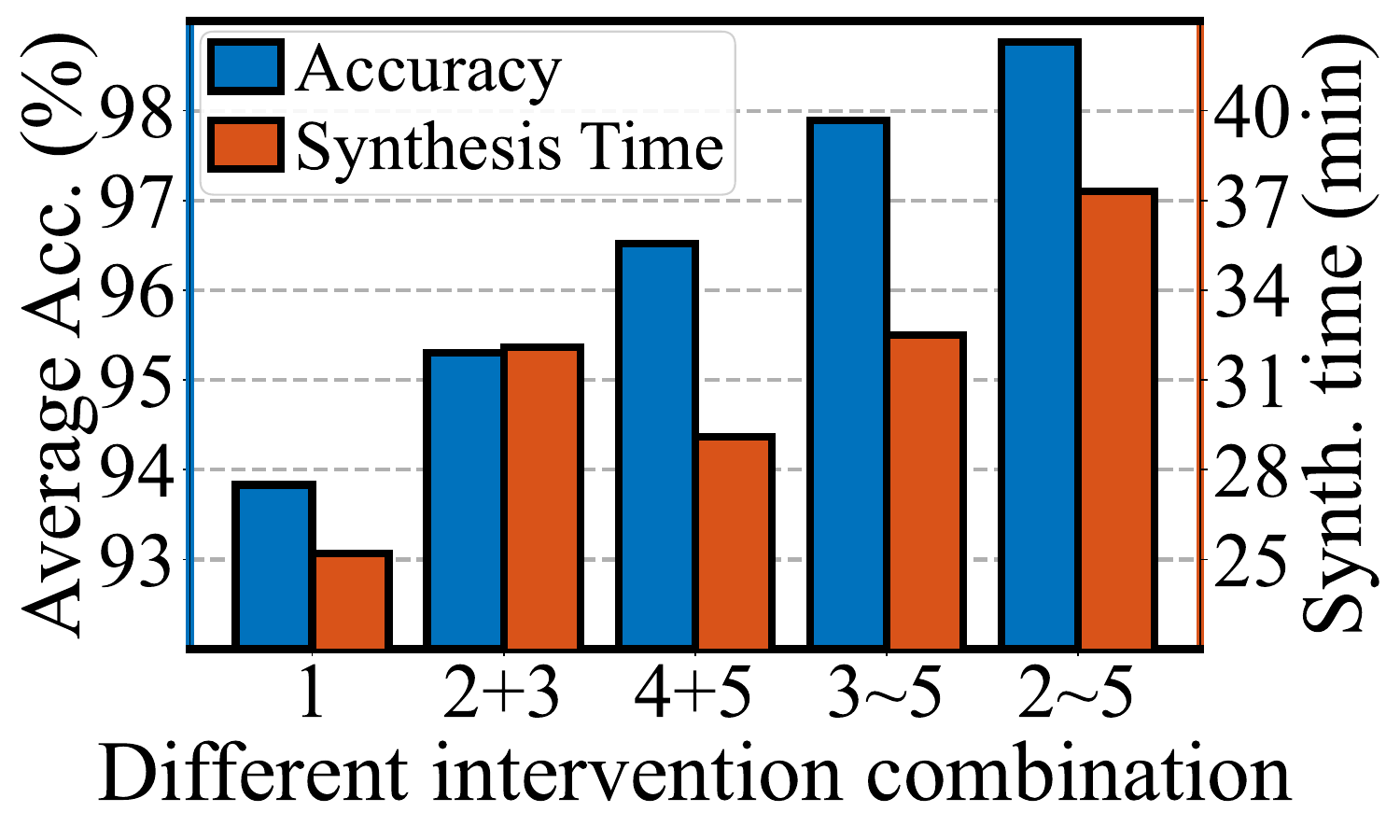}
        }
        \caption{Different levels of user intervention.}
    \end{minipage}
    \begin{minipage}[t]{0.49\textwidth}
        \setlength{\abovecaptionskip}{-4pt}
        \subfigtopskip=-6pt
        \subfigcapskip=-6pt
            \centering
            \subfigure[GPT-4]{
                \label{fig:iteration_gpt4}
                \includegraphics[width=0.45\textwidth]{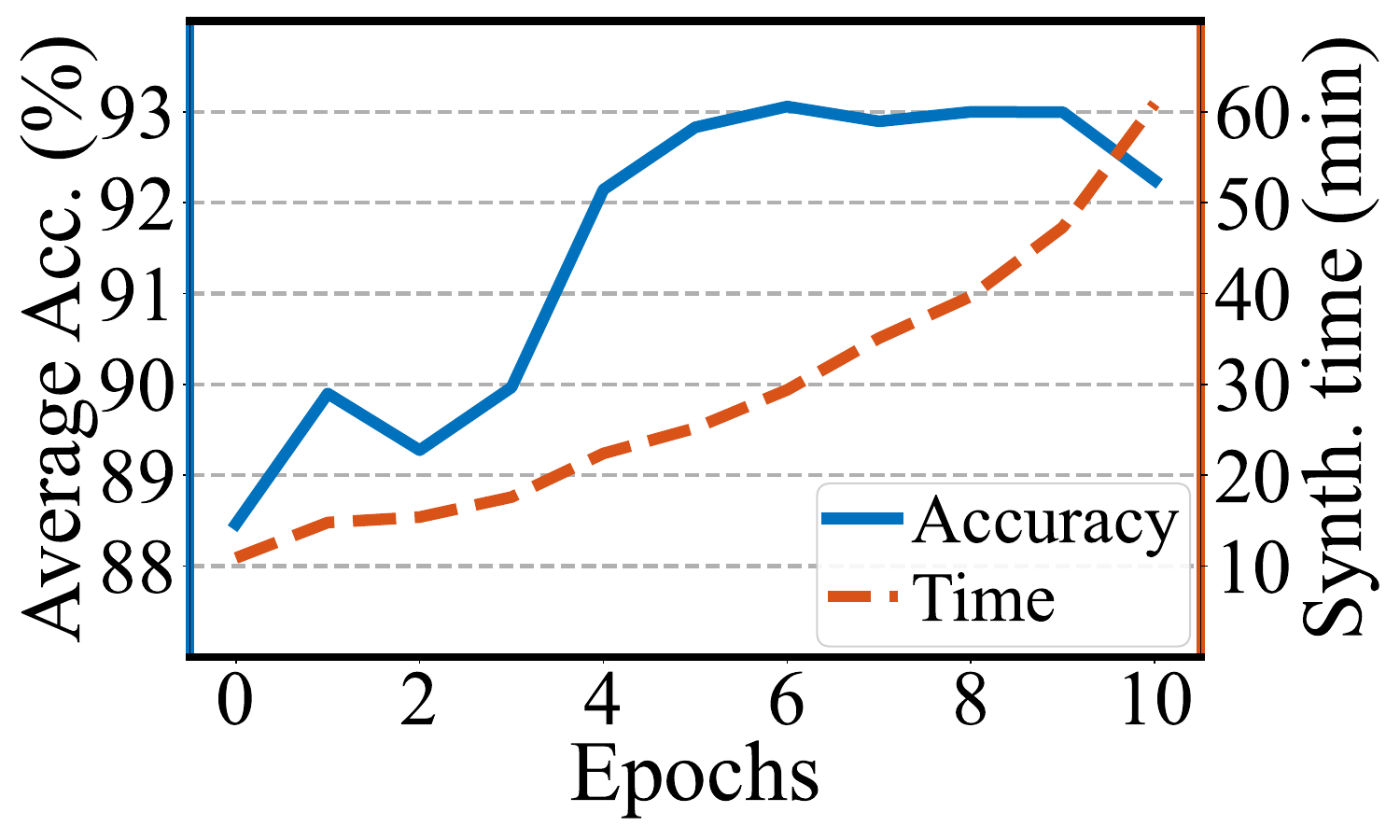}
            }
            \centering
            \subfigure[GPT-3.5]{
                \label{fig:iteration_gpt3}
                \includegraphics[width=0.45\textwidth]{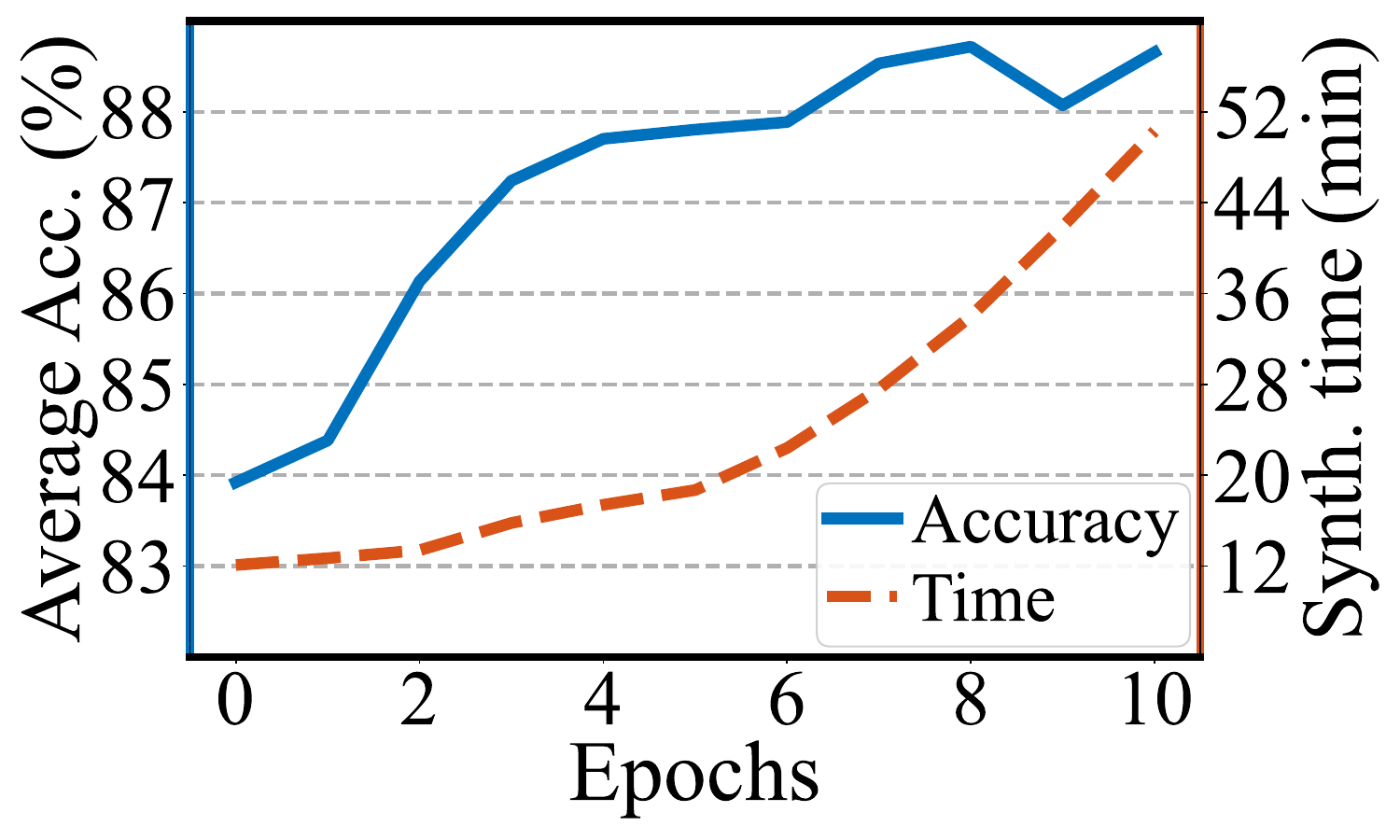}
            }
            \caption{Different numbers of iteration.}
            \label{fig:iteration}
    \end{minipage}
    \vspace{-18pt}
\end{figure*}

For multimodal HAR, the input instruction (A1) includes the basic information of the task, \ie, the task target, the dataset specifications, and the output format. Based on that, we create two additional variations: one with a GPU-memory-constrained requirement (A2) and another with a high accuracy requirement (A3). We then feed the instructions into \textit{AutoIOT} and measure the accuracy and inference time of the synthesized programs, with results shown in Fig.~\ref{fig:overall_multimodal}. By analyzing the three synthesized different programs, we observe that: 1) All the synthesized programs adopt a similar workflow as the baseline system, \ie, they first construct three encoders to extract effective features from the three modalities, then concatenate these features and feed them into a classifier for activity recognition. This implies that benefiting from our CoT-based problem-solving paradigm, \textit{AutoIOT} recognizes the workflow and architecture as effective and standard for handling multimodal data-related tasks, which is consistent with most of the existing methods \cite{ouyang2023harmony}. 2) \textit{AutoIOT} can adjust the generated code to fulfill different requirements. The second program consumes less memory than others due to the resource constraint requirement, resulting in lower accuracy but reduced inference time (Fig.~\ref{fig:memory_time}). On the other hand, the third program adopts a more complex and larger model architecture, requiring more GPU memory and incurring a longer inference time. Such differences validate the capabilities of \textit{AutoIOT} in accurately understanding and processing natural language-based user requirements. These observations further demonstrate the effectiveness of \textit{AutoIOT} in ensuring the correctness of user requirement understanding and the generated code, benefiting from our automatic self-improvement component.



\vspace{-5pt}
\subsubsection{Communication cost. \& wall-clock time}
We select heartbeat detection as an example and measure the total communication cost with wall-clock execution time of \textit{AutoIOT} and direct LLM inference as done in Penetrative AI \cite{10.1145/3638550.3641130}. Specifically, ECG data is first down-sampled and segmented into multiple windows and then embedded into the prompt for LLMs' inference. Experiment results show that \textit{AutoIOT} requires 8MB of network traffic mainly for prompt transmissions, while \cite{10.1145/3638550.3641130} consumes more than 50MB mainly for sensor data transmissions. Besides, \textit{AutoIOT} takes 25 minutes to complete the task with a dramatic reduction in inference time compared to \cite{10.1145/3638550.3641130}, which needs to send and process all windowed signals with the remote LLM serving.

\vspace{-10pt}
\subsection{Sensitivity Analysis}
\vspace{-2pt}

\noindent\textbf{Different levels of user intervention.} To show how \textit{automated program synthesis} improves user experiences, we evaluate the \textit{AutoIOT}'s performance under five different levels of user intervention: 1) No intervention; 2) Intervention with user-provided domain knowledge; 3) Intervention with user-specified algorithms for program synthesis; 4) Intervention with user-based debugging; 5) Intervention with user-decided algorithm modification for code improvement.
Fig.~\ref{fig:intervention_single} shows the performance of \textit{AutoIOT} with different levels of user intervention. When the user manually instructs the LLM to generate code according to specific hand-picked algorithms (\eg, designed by experts or research papers), the average accuracy can be improved. This user-in-the-loop process becomes particularly advantageous when users have a higher level of domain expertise in AIoT, enabling them to design or select more advanced and robust algorithms. 
But it leads to increased synthesis time as the LLM has to revise outputs until the user is satisfied. Fig.~\ref{fig:intervention_combined} shows the performance with different user intervention combinations. We see that increased user involvement in the program synthesis process correlates with higher accuracy. However, this heightened engagement leads to significantly longer synthesis time and extra user overhead.
Note that \textit{AutoIOT} may not always be able to fix bugs and finish program synthesis tasks. In this case, user intervention with minimum effort is still a must. Thus, \textit{AutoIOT} allows users to provide detailed instructions necessary for program synthesis by the LLMs.


\begin{figure}[t]
\vspace{8pt}
\setlength{\abovecaptionskip}{-4pt}
\subfigtopskip=-6pt
\subfigcapskip=-6pt
    \centering
    \subfigure[Average accuracy \& MAE]{
        \label{fig:llm_acc}
        \includegraphics[width=0.225\textwidth]{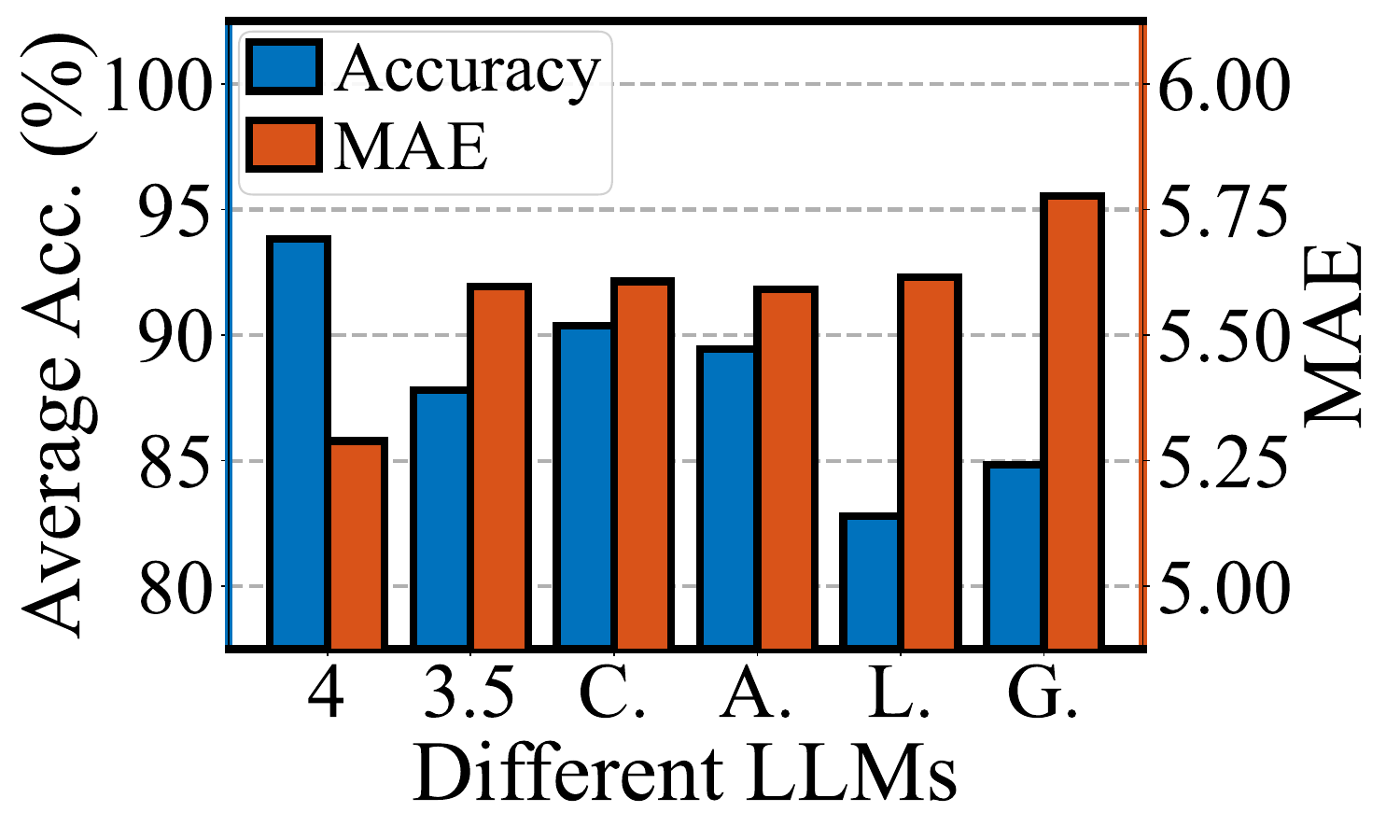}
    }
    \centering
    \subfigure[Wall-clock time \& network]{
        \label{fig:llm_time}
        \includegraphics[width=0.225\textwidth]{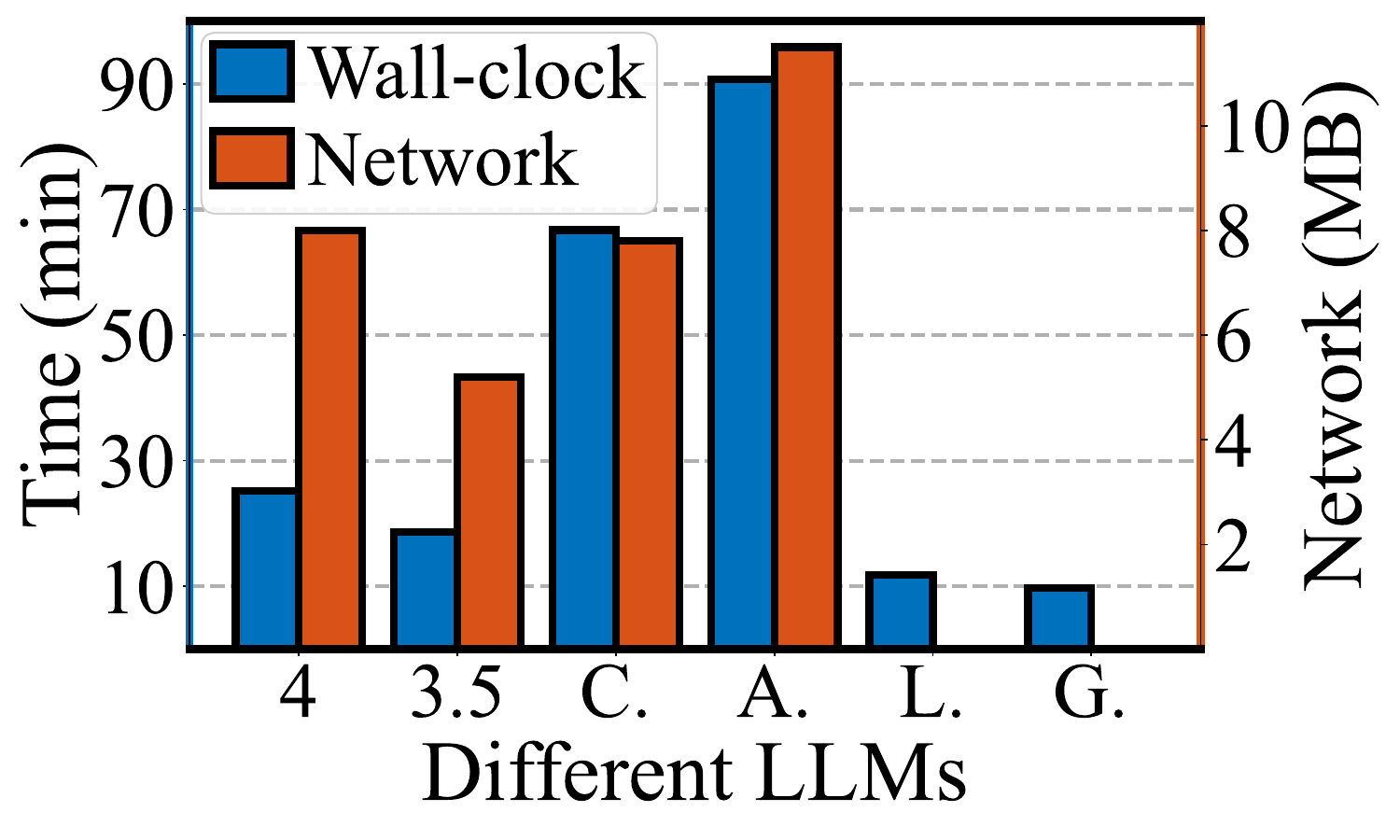}
    }
    \caption{Different LLMs. (4 for GPT-4, 3.5 for GPT-3.5, C. for Cohere, A. for Anthropic Claude 2, L. for Llama2-7b, and G. for Gemma-7b.)}
    \vspace{-18pt}
\end{figure}

\begin{figure*}[t]
\vspace{-8pt}
\setlength{\abovecaptionskip}{-4pt}
\subfigtopskip=-6pt
\subfigcapskip=-6pt
    \centering
    \subfigure[Single component ablation]{
        \label{fig:ablation_single}
        \includegraphics[width=0.35\textwidth]{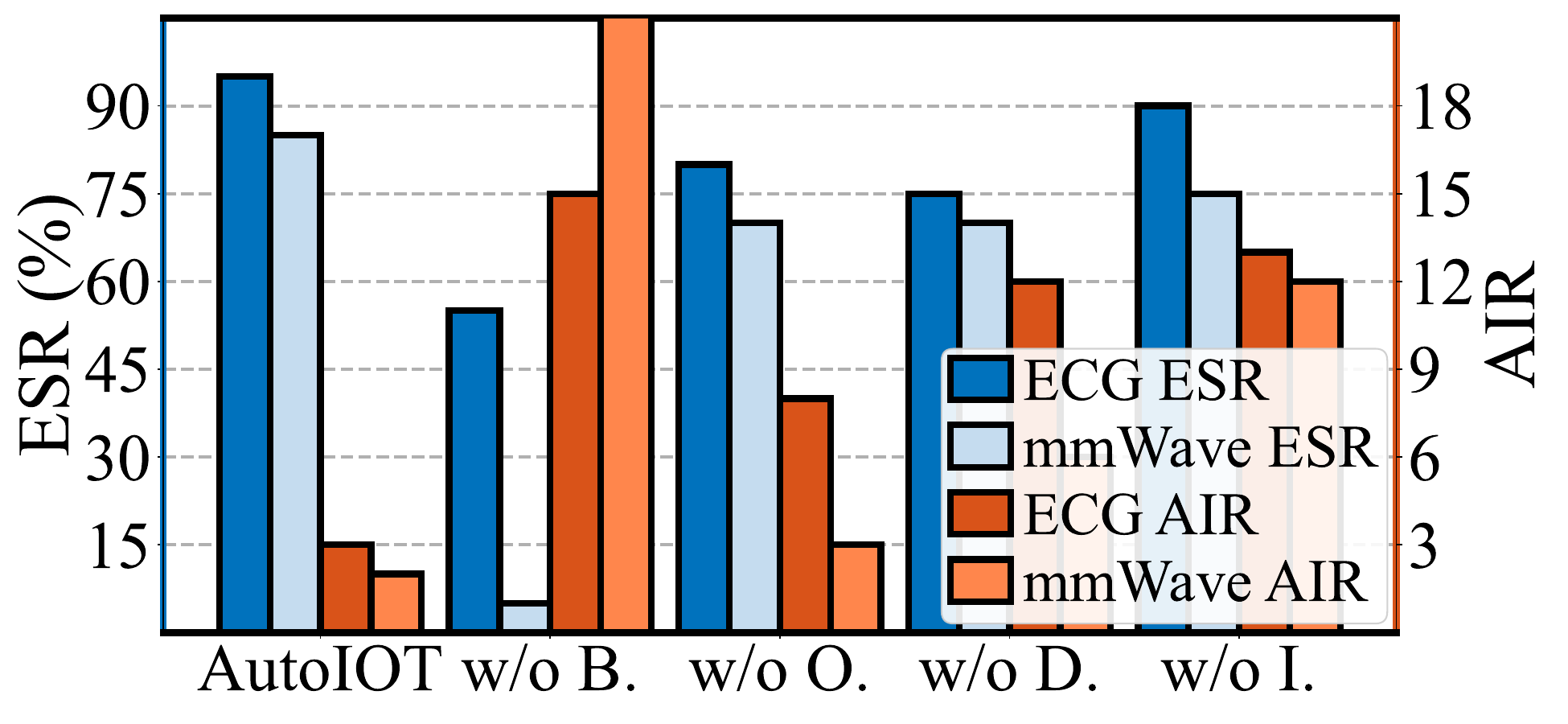}
    }
    \centering
    \subfigure[Combined component ablation]{
        \label{fig:ablation_combine}
        \includegraphics[width=0.35\textwidth]{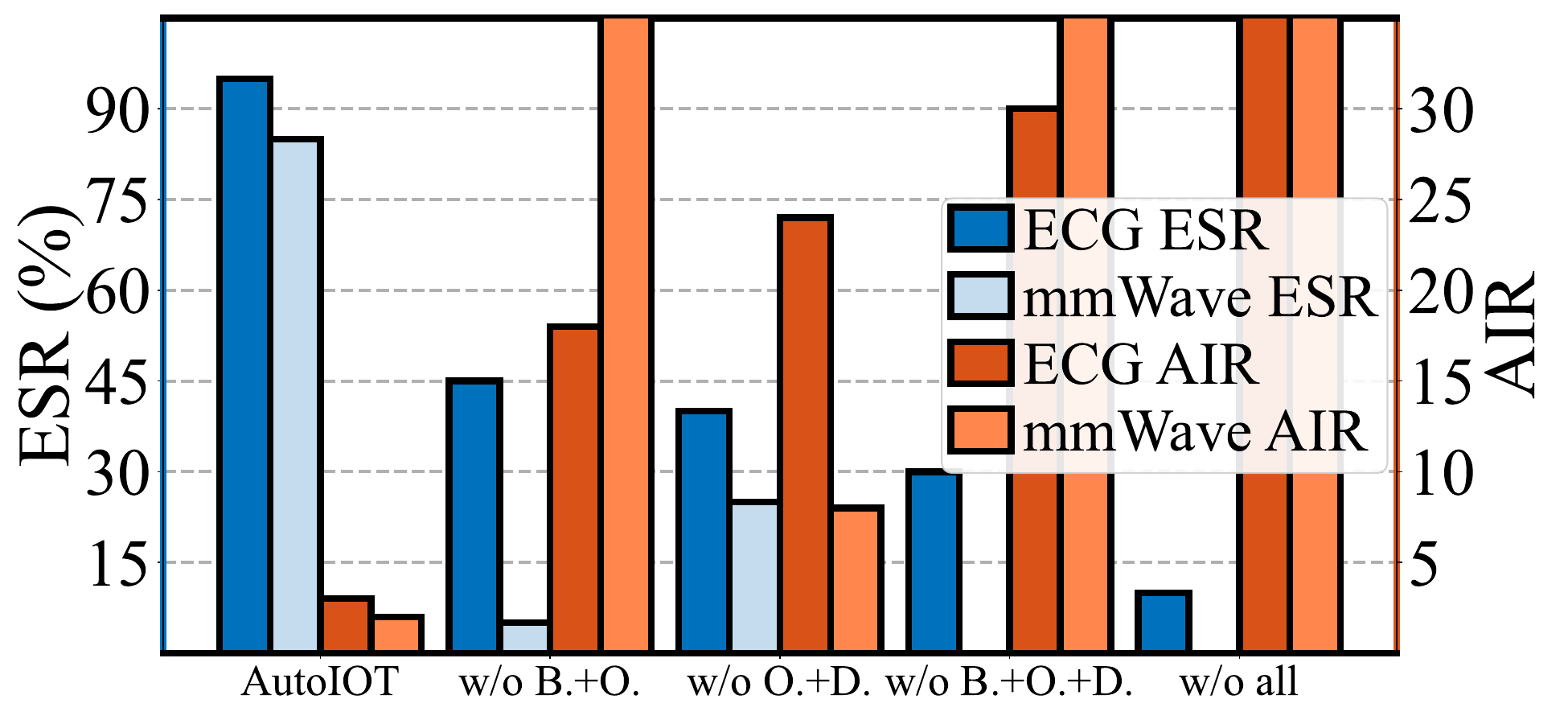}
    }
    \centering
    \subfigure[User study (subjective)]{
        \label{fig:user_study_subjective}
        \includegraphics[width=0.18\textwidth]{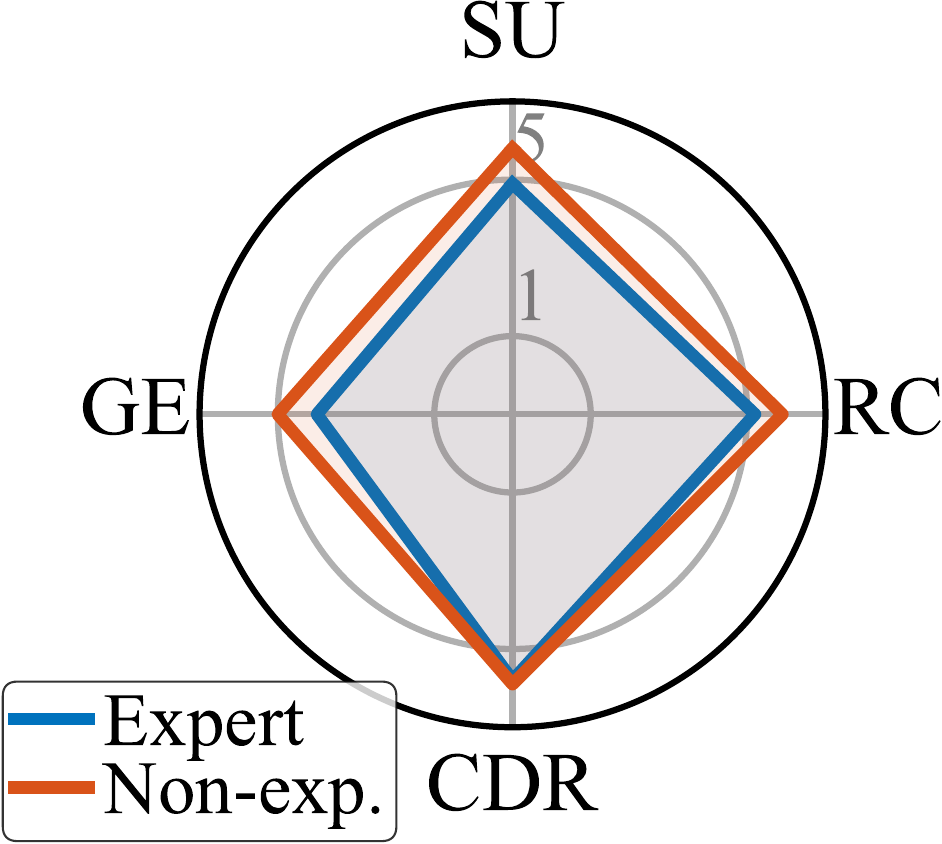}
    }
    \caption{(a) \& (b): Ablation study. B. for background knowledge retrieval, O. for algorithm outline generation, D. for detailed design generation, and I. for code improvement. (c): User study on subjective metrics.}
    \label{fig:ablation}
    \vspace{-18pt}
\end{figure*}

\noindent\textbf{Different numbers of improvement iterations.}
We vary the number of epochs for code optimization from 0 to 10 and evaluate the impact on the synthesized programs. As shown in Fig.~\ref{fig:iteration}, with more improvement epochs, the accuracies/synthesis time of the GPT-3.5/-4 generated programs gradually increase. However, after around 5 epochs, the marginal gain of average accuracy starts to diminish while the synthesis time increases dramatically. 
This is because, with longer conversation history, LLMs may fail to recall past context information and tend to generate inconsistent responses \cite{wang2023recursively}. Therefore, we empirically set the epoch number to five.

\begin{figure}[t]
\vspace{8pt}
\setlength{\abovecaptionskip}{-4pt}
\subfigtopskip=-6pt
\subfigcapskip=-6pt
    \centering
    \subfigure[Task accuracy]{
    \label{fig:user_study_accuracy}
        \includegraphics[width=0.225\textwidth]{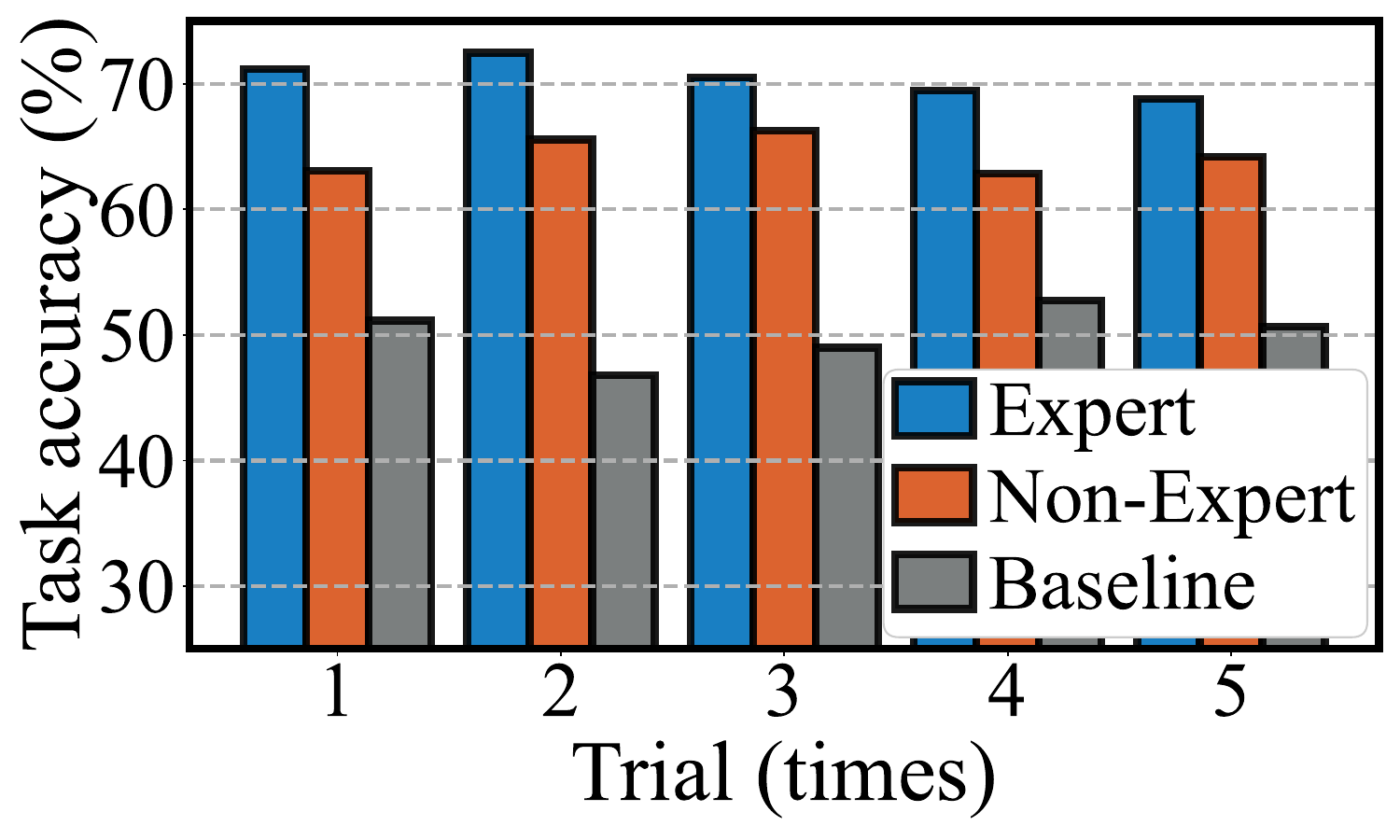}
    }
    \centering
    \subfigure[Code correctness verification]{
        \label{fig:user_study_correctness}
        \includegraphics[width=0.225\textwidth]{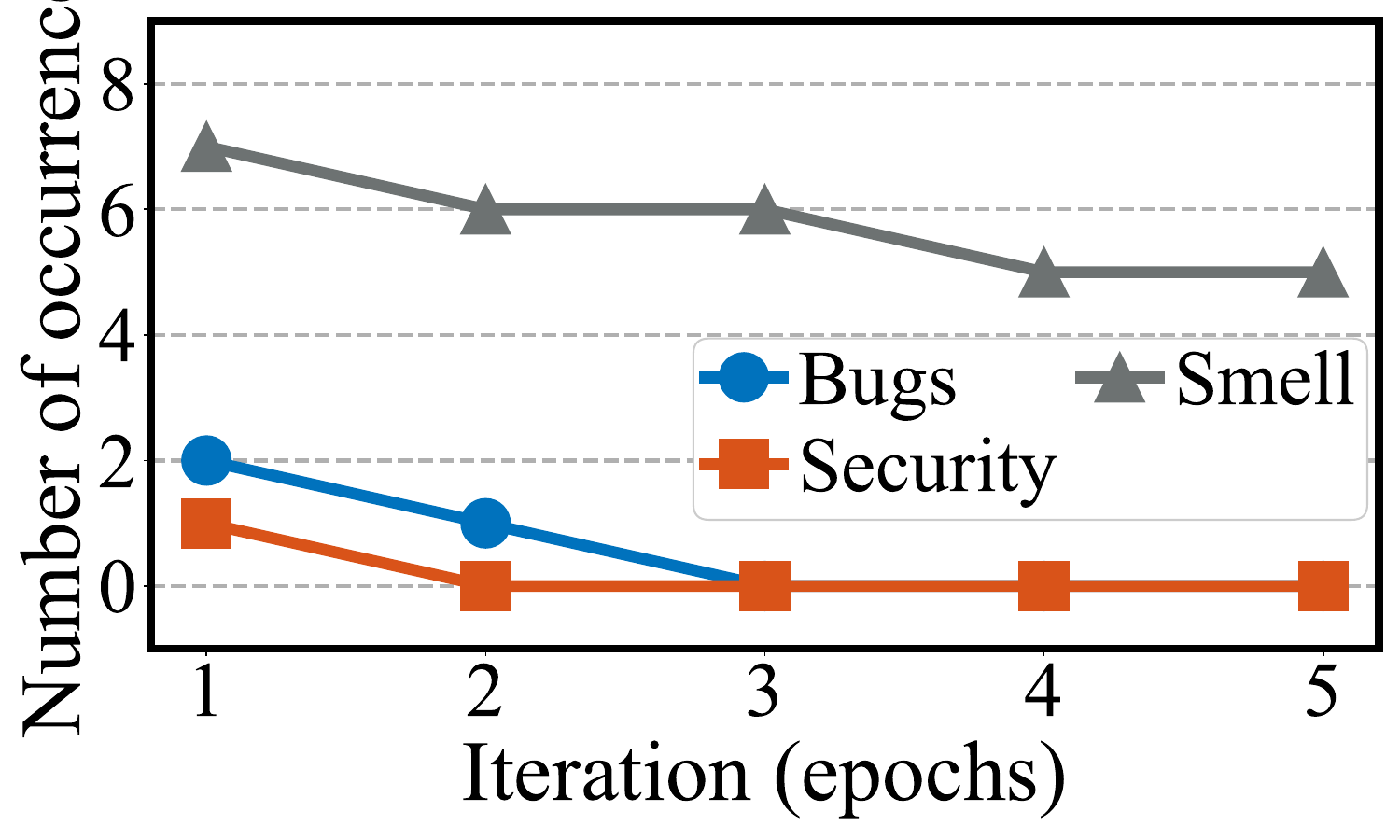}
    }
    \caption{User study (Objective Evaluation).}
    \vspace{-18pt}
\end{figure}

\noindent\textbf{Impact of Different LLMs.} We select the following LLMs for comparison: GPT-4 \cite{achiam2023gpt}, GPT-3.5 \cite{brown2020language}, Llama2-7b \cite{touvron2023llama}, Cohere \cite{Cohere2023}, Claude 2 \cite{AnthropicClaudeWeb2023}, and Gemma-7b \cite{gemma2024}. 
Llama2 and Gemma are locally deployed in our lab.
We select R-peak detection as an example with the Christov algorithm as the baseline. Fig.~\ref{fig:llm_acc} shows that GPT-4 performs the best. 
Given the knowledge retrieved by the same tools, LLMs still need strong language understanding and reasoning capabilities to comprehend AIoT tasks and synthesize programs. 
Experiment results indicate that GPT-4 might have superior performance in language understanding and reasoning capability for this specific task. 
Although Llama2-7b and Gemma-7b achieve relatively lower accuracy, these two local models offer much faster response speeds.



\vspace{-8pt}
\subsection{Ablation Study}
\label{sec:AblationStudy}
In the ablation study, we use two metrics to evaluate the code quality: 1) \textit{Execution success rate} (ESR): the proportion of the code that can be executed successfully for the first time. 2) \textit{Average iteration round} (AIR): the average number of improvement iterations required to achieve 80\% accuracy.

\noindent\textbf{Background knowledge retrieval.}
To explore the influence of the \textit{background knowledge retrieval} module, we disable the web search tool and the knowledge database. We then instruct the LLM to synthesize 20 different programs with no user intervention. Fig.~\ref{fig:ablation_single} shows the ESR and AIR of two AIoT tasks. We see that the ESR of heartbeat detection drops slightly while the ESR of mmWave-based HAR exhibits a significant drop. This is because the mmWave-based HAR uses a newly published dataset, which has not been seen by the LLM. Therefore, the LLM does not know the dimensionality of the dataset and only randomly configures the hyper-parameters of the neural network. Additionally, both applications require larger numbers of iterations to improve the accuracy of synthesized programs. We note that mmWave-based HAR even fails to achieve the expected accuracy (thus, its AIR is marked as infinite). A similar phenomenon is also observed in Fig.~\ref{fig:ablation_combine}. The experiment results indicate that the background knowledge retrieval module plays a pivotal role for the LLM to retrieve up-to-date domain knowledge to augment the program synthesis process.  
\noindent\textbf{Chain-of-thought.}
We evaluate the contribution of the algorithm outline generation step and the detailed design generation step during the CoT prompting, respectively. As shown in Fig.~\ref{fig:ablation_single}, when only one step is enabled, 
we observe a slight drop in ESR and an increase in AIR for both applications. When we disable both steps, the ESR drops significantly as shown in Fig.~\ref{fig:ablation_combine}. 
Without the explicit guidance specified in the two steps, the LLM cannot synthesize executable programs and presents null functions with placeholders. 
Therefore, the CoT method with detailed instructions emulating the software development lifecycle plays a crucial role in helping LLMs synthesize executable programs.


\noindent\textbf{Code improvement.}
The \textit{code improvement} module contains automated debugging and code optimization. Since automated debugging is essential to ensure the executability of synthesized programs, we only conduct an ablation study on the code optimization step. With the program after the debugging step, we evaluate the \textit{code improvement} module by directly instructing the LLM to modify the program over several iterations without providing the compiler or interpreter feedback. As shown in Fig.~\ref{fig:ablation_single}, without the \textit{code improvement} module, the ESR is almost unaffected while the AIR increases significantly. 
This is because generating an executable program with no syntax errors is seldom influenced by this module.
However, without feedback from the compiler or interpreter, more iterations are required since the LLM does not know which step of the algorithm should be modified or improved. Besides, due to the laziness of LLMs \cite{liu2024lost}, synthesized programs tend to adopt simple and popular algorithms. This necessitates the \textit{code improvement} module, which progressively directs the LLM to explore more advanced algorithms to improve the synthesized program.  



\vspace{-8pt}
\subsection{User Study}
\vspace{-2pt}

To investigate the utilities of \textit{AutoIOT}, we conduct a user study (N=20) by inviting 5 expert and 15 non-expert users, whose detailed background information is listed in Table~\ref{tab:user_information}. The expert users are PhD students and professors with work or research experience in the IoT field and have developed many IoT applications. We select human activity recognition using RFID data (XRF55 dataset \cite{wang2024xrf55}) as the IoT application, where a 1D Conv-based ResNet18 is the baseline.

\begin{table}[t]
\vspace{-15pt}
\centering
\scriptsize
\setlength{\abovecaptionskip}{-1pt}
\caption{Participant information of user study (N=20)}
\label{tab:user_information}
\begin{tabular}{ll}
\toprule[1pt]
\textbf{Category}                             & \textbf{Background Information}                                                                 \\ \toprule[1pt]
\textbf{Gender}                               & Female (45\%), Male (40\%), Prefer not to say (15\%)\\ 
\textbf{Age}                                  & Under 18 (10\%), 18-30 (75\%), 30-39 (10\%), 40 and older (5\%)           \\ 
\textbf{Education}                      & Bachelor (20\%), Master (15\%), Doctoral (60\%), Others (5\%) \\ 
\textbf{English}     & Beginner (5\%), Intermediate (25\%), Advanced (60\%), Fluent (10\%)                              \\ 
\textbf{Expertise} & Expert (25\%), Non-expert (75\%)               \\ \bottomrule[1pt]
\end{tabular}
\vspace{-18pt}
\end{table}

\noindent\textbf{Objective Evaluation.}
We first repeatedly measure the average task accuracy (classification accuracy) after executing the synthesized programs, with the results shown in Fig.~\ref{fig:user_study_accuracy}.
We see that the programs synthesized by the two groups of users outperform the baseline across multiple trials. Besides, the programs synthesized for experts typically perform better than those for non-experts. The main reason is that experts tend to provide more information in the specifications (\eg, the dataset format, the training workflow). Consequently, \textit{AutoIOT} can synthesize programs with more advanced algorithms and detailed specifications for expert users. Next, we use SonarQube \cite{sonar} to verify the correctness of the generated code, including bug/logic errors, security issues, and code smells \cite{sonarsource} after every improvement iteration. Code smells are not bugs but bad coding styles (\eg, variable name mismatching regular expression) or potential weaknesses (\eg, package version incompatibility). From Fig~\ref{fig:user_study_correctness}, we observe that several bugs and one security issue present initially are ultimately fixed by \textit{AutoIOT}. \textit{AutoIOT} may not be able to address all code smells, as they are closely related to coding styles \cite{liu2019deep}. Applying code smell correction to the retrieved algorithms and programs can be promising for \textit{AutoIOT} to iteratively detect and fix code smell-related issues.

\noindent\textbf{Subjective Measurement.}
We ask the users to execute the synthesized programs and rate \textit{AutoIOT} based on four subjective metrics: 1) \textit{System Utility} \textbf{(SU)} measures the user's overall satisfaction with \textit{AutoIOT}'s performance; 2) \textit{Requirement Coverage} \textbf{(RC)} evaluates how well the user requirements are fulfilled by \textit{AutoIOT}; 3) \textit{Code \& Documentation Readability} \textbf{(CDR)} measures the clarity and structure of the code and documentation; 4) \textit{Generation Efficiency} \textbf{(GE)} accesses how acceptable the waiting time is for synthesizing the final program. All the above metrics are rated by the users on a scale from 1 (not at all) to 6 (more than expected). As shown in Fig.~\ref{fig:user_study_subjective}, the average GE of both user groups reaches 4.5, indicating that the waiting time for \textit{AutoIOT} to synthesize the final program is acceptable for most users. Additionally, we find that non-experts tend to give higher scores (SU, RC, and CDR) than experts. Further examination reveals that non-experts tend to under-specify their requirements.
Surprisingly, LLMs can sometimes provide comprehensive responses to meet their requirements. This ability is rooted in LLMs' extensive training on diverse datasets and retrieved online information, enabling them to infer and bridge the gaps with relevant information \cite{tamkin2021understanding, floridi2020gpt}.

\begin{figure}[t]
\vspace{-8pt}
\setlength{\abovecaptionskip}{-4pt}
\subfigtopskip=-5pt
\subfigcapskip=-5pt
    \centering
    \subfigure[Runtime efficiency optimization]{
        \label{fig:runtime_efficiency}
        \includegraphics[width=0.234\textwidth]{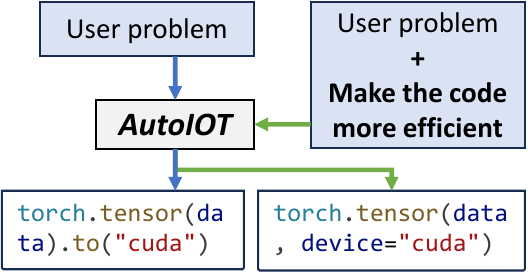}
    }
    \centering
    \subfigure[Tailored for target platform]{
        \label{fig:target_platform}
        \includegraphics[width=0.216\textwidth]{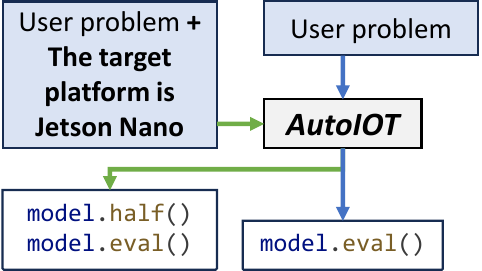}
    }
    \caption{Further experiments.}
    \vspace{-18pt}
\end{figure} 

\vspace{-8pt}
\section{Discussion}
\vspace{-2pt}
\textbf{Runtime Efficiency.} \textit{AutoIOT} focuses on generating functionally correct code for IoT data processing. While the runtime efficiency of the programs is not optimized, the users can specify the extra requirements in the prompts. Fig.~\ref{fig:runtime_efficiency} shows an example that given the user specification, the generated code adopts CUDA optimization and thereby achieves similar runtime efficiency compared to expert-optimized programs. This is because \textit{AutoIOT} can retrieve and learn from hand-crafted optimizations via various online sources.

\noindent\textbf{Workflow Generalizability.} 
Existing IoT applications can be primarily categorized into four types based on the specific stages of the IoT workflow they address: 1) data collection applications rely on sensors (\eg, IMU and radar) to gather raw information from the environment; 2) data transmission applications enable seamless communication across IoT networks to cloud or edge systems; 3) data processing applications typically adopt advanced algorithms or AI models to analyze and process IoT data; 4) decision-making and actuation applications parse the processed data to perform task automation or actuator control.
In this paper, \textit{AutoIOT} is designed primarily for IoT data processing tasks by offering end-to-end solutions with executable programs. The prompts designed in \textit{AutoIOT} are not limited to specific data patterns (time-series or high-dimensional) and processing pipelines (sequentially or parallelly). For other IoT application workflows, users can specify the requirements in the prompt.
For example, users can first instruct \textit{AutoIOT} to develop a WiFi data collection application. Then, the synthesized program can be deployed on WiFi-related devices to capture WiFi data. Next, users can further request \textit{AutoIOT} to process and analyze the collected WiFi data to perform HAR or other tasks. Moreover, some IoT applications may require executing multiple programs simultaneously for cross-device communication or synchronization. To tackle such applications, a possible solution can be deploying multiple \textit{AutoIOT} agents with advanced collaboration mechanisms for enhanced cross-device and cross-program interaction.
We leave the comprehensive generalization to other types of IoT tasks beyond pipelined data processing for future work.

\noindent\textbf{Real-World Deployment.} We show that \textit{AutoIOT} can synthesize functionally correct IoT programs in Python with data processing accuracy as the main performance metric. 
\textit{AutoIOT} is not limited to specific target IoT platforms. For example, regarding those resource-constrained MCU-class devices \cite{shen2024fedconv, wang2024latte}, various practical requirements (\eg, power management and network protocols) of the IoT device can be factored into \textit{AutoIOT}'s prompts for program synthesis. Fig.~\ref{fig:target_platform} shows an example where a user explicitly specifies Jetson Nano, which has limited GPU resources, as the target platform. \textit{AutoIOT} will then adopt half-precision training and inference for the AI model to save GPU memory. Moreover, users can even provide \textit{AutoIOT} with handbooks of some dedicated IoT devices for reference.


\noindent\textbf{Memorization Issue.} In \textit{AutoIOT}, sometimes the synthesized programs' performance cannot be improved, even after multiple improvement iterations. This is because LLMs may forget the previous context, resulting in inconsistent code improvement suggestions during iterations \cite{chen2023forgetful}. In such cases, our user-in-the-loop optimization allows users to optionally provide instructions that can help the LLM improve the synthesized programs. For instance, users can explicitly instruct the LLM to use the Pan-Tompkins algorithm for ECG data processing. Additionally, we can further upgrade \textit{AutoIOT} by adopting existing memorization enhancement methods, such as context compression via RAG \cite{shi2024compressing} and iterative summarization \cite{sun2024prompt}. By integrating these orthogonal approaches with \textit{AutoIOT}, each self-improvement iteration can contribute more positively to the performance of the generated code with enhanced context consistency.

\noindent\textbf{Privacy Concerns.} In \textit{AutoIOT}, only user requirements are transmitted to the cloud LLM for processing. We proactively instruct \textit{AutoIOT} to treat user configuration data (\eg, the local file path) as program input. As such, during local execution, users can directly input the private configurations in the console (\S~\ref{sec:automation}), rather than pre-defining them in the prompts for cloud LLMs. To further mitigate privacy concerns, one possible solution can be deploying a local LLM to handle all code generation and debugging tasks \cite{shen2024iotcoder, shen2025gpiot}.



\vspace{-8pt}
\section{Related Work}

\vspace{-2pt}
\textbf{Integration of LLMs with AIoT.} Existing integration methods have two main types: 1) Prompt-based methods embed raw sensor data into tailored prompts and instruct LLMs to perform various AIoT tasks. HARGPT \cite{ji2024hargpt} and LLMSense \cite{ouyang2024llmsense} embed textualized sensor data into prompts to show the proficiency of LLMs in comprehending IoT sensor data. They require transmitting raw sensor data to LLM servers, suffering similar issues as Penetrative AI. 2) Fine-tuning-based methods retrain LLMs with labeled datasets containing sensor data inference examples. LLM4TS \cite{chang2024llm4ts} fine-tunes LLMs using sensor data with labels for time-series data prediction. However, these works demand high compute and memory resources.
In contrast, \textit{AutoIOT} explores a new approach to automatically synthesizing programs for AIoT applications without extra overheads. 


\noindent\textbf{Code LLMs}.
Recent advances in code LLMs \cite{roziere2023code, guo2024deepseek} have demonstrated the potential to revolutionize software development. They can produce functionally correct code across various programming languages and input compiler output into LLMs for debugging and program refining \cite{zhong2024ldb, duan2023leveraging}. However, they require high computing resources with carefully selected datasets (StarCoder \cite{li2023starcoder} needs an 815GB dataset and 512 A100 80GB GPUs). Worse still, they are unaware of the latest advances in highly specialized domains and cannot generate comprehensive solutions for complex IoT tasks. \textit{AutoIOT} draws strength from these works and addresses specific technical challenges in IoT program synthesis, which require ever-evolving domain knowledge that the above LLMs have not yet assimilated. Table~\ref{tb:comparison} compares \textit{AutoIOT} with SOTA code LLMs. LLM-based code generation synthesizes functionally correct programs for a well-defined module, whereas LLM-based task automation generates a complete automation chain from requirements to solutions. \textit{AutoIOT} lies at the intersection of these two approaches, with the objective of code generation and task automation. Additionally, RAG and automated code improvement jointly improve the code generation process in \textit{AutoIOT}.

\begin{table}[t]
\vspace{-10pt}
\scriptsize
\setlength{\abovecaptionskip}{4pt}
\caption{\textit{AutoIOT} vs SOTA code LLMs}
\label{tb:comparison}
\centering
\begin{tabular}{ccccccc} 
\toprule[1pt]
\textbf{Type}                        & \textbf{Name}           & \textbf{\makecell{Mod.\\ Gen.}}          & \textbf{\makecell{Sys.\\Gen.}}         & \textbf{RAG}                        & \textbf{Auto.}                  & \textbf{\makecell{Auto Debug\\\& Impro.}}      \\ 
\toprule[1pt]
\multirow{3}{*}{\textbf{\makecell{Code\\Gen.}}} & \makecell{DeepSeek\\Coder \cite{guo2024deepseek}} & \ding{51} & \ding{55} & \ding{55} & \ding{55} & \ding{55}  \\
                            & CodeLlama \cite{roziere2023code}     & \ding{51} & \ding{55} & \ding{55} & \ding{55} & \ding{55}  \\
                            & WizardCoder \cite{luo2023wizardcoder}    & \ding{51} & \ding{55} & \ding{55} & \ding{55} & \ding{55}  \\ 
\toprule[1pt]
\multirow{2}{*}{\textbf{\makecell{Task\\Auto.}}}  & AutoGPT \cite{autogpt}        & \ding{51} & \ding{51} & \ding{55} & \ding{51} & \ding{55}  \\
                            & MetaGPT \cite{hong2023metagpt}         & \ding{51} & \ding{51} & \ding{55} & \ding{51} & \ding{55}  \\ 
\toprule[1pt]
\multicolumn{2}{c}{ \textbf{\textit{AutoIOT}}}                & \ding{51} & \ding{51} & \ding{51} & \ding{51} & \ding{51}  \\
\bottomrule[1pt]
\end{tabular}
\vspace{-10pt}
\end{table}



\vspace{-8pt}
\section{Conclusion}
\vspace{-2pt}
We propose \textit{AutoIOT}, an LLM-driven automated natural language programming system for AIoT applications. Our system features three novel technical modules: background knowledge retrieval, automated program synthesis, and code improvement, transforming natural language descriptions into executable programs.
Our experiments demonstrate the competitive performance of \textit{AutoIOT} in synthesizing programs for a variety of AIoT applications, with comparable performance in challenging AIoT tasks and sometimes outperforming some representative baselines. This showcases the strong potential of exploiting the embedded common knowledge of LLMs to evolve AIoT application development. 

\vspace{-5pt}
\begin{acks}
\vspace{-2pt}
We sincerely thank our shepherd -- Mi Zhang, and anonymous reviewers for their constructive comments and invaluable suggestions that helped improve this paper. This work is supported by Hong Kong GRF Grant No. 15211924, 15206123, and 16204224. Yuanqing Zheng and Mo Li are the corresponding authors.
\end{acks}

\clearpage
\balance
\bibliographystyle{ACM-Reference-Format}

\end{sloppypar}
\end{document}